\documentclass[runningheads]{llncs}

\usepackage{eccv}

\usepackage{eccvabbrv}

\usepackage{graphicx}
\usepackage{booktabs}
\usepackage{adjustbox}
\usepackage{multirow}
\usepackage{comment}
\usepackage{dsfont}
\usepackage[dvipsnames]{xcolor}

\makeatletter
\renewcommand{\paragraph}{%
  \@startsection{paragraph}{4}%
  {\z@}{0.25em}{-1em}%
  {\normalfont\normalsize\bfseries}%
}
\makeatother

\usepackage{xspace}

\makeatletter
\DeclareRobustCommand\onedot{\futurelet\@let@token\@onedot}
\def\@onedot{\ifx\@let@token.\else.\null\fi\xspace}
\def\eg{\emph{e.g}\onedot} 
\def\ie{\emph{i.e}\onedot} 
\def\cf{\emph{c.f}\onedot} 
\def\etc{\emph{etc}\onedot}

\makeatother

\usepackage{amssymb}
\usepackage{pifont}
\newcommand{\cmark}{\ding{51}}%
\newcommand{\xmark}{\ding{55}}%

\newlength{\faHeight}
\newcommand{\faBird}{\includegraphics[height=\faHeight]{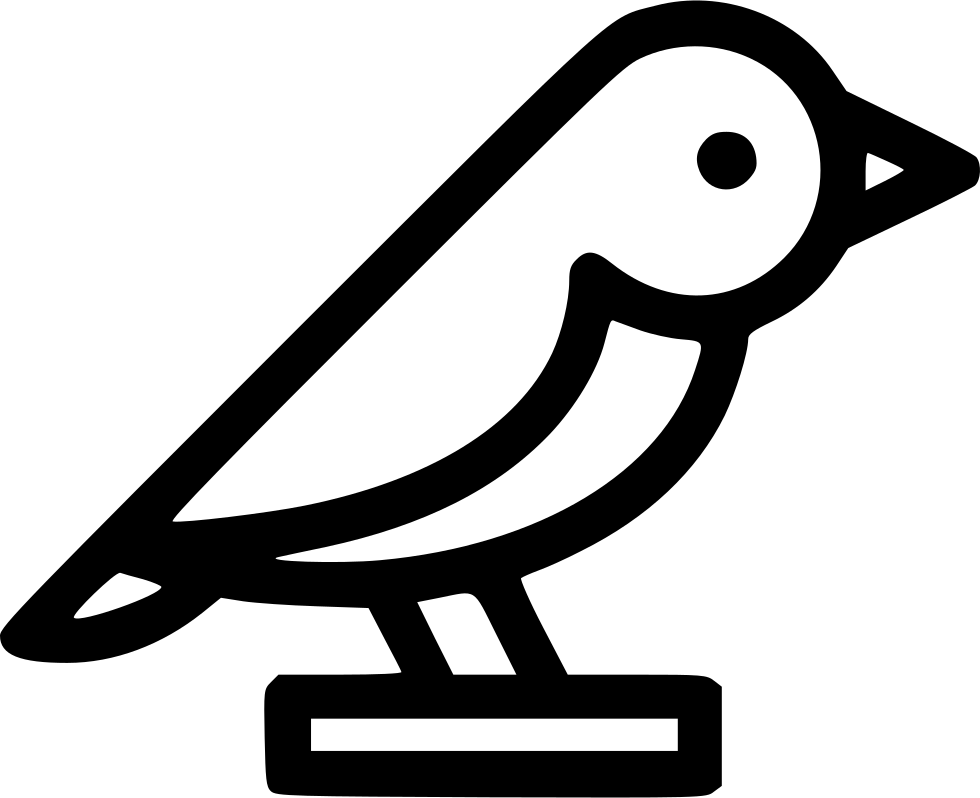}}
\newcommand{\faBus}{\includegraphics[height=\faHeight]{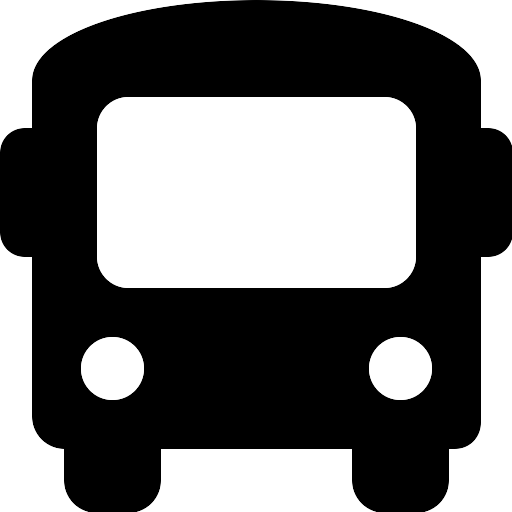}}
\newcommand{\faCar}{\includegraphics[height=\faHeight]{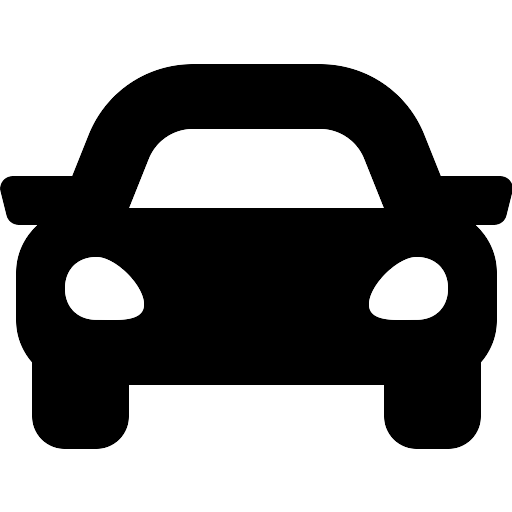}}
\newcommand{\faCat}{\includegraphics[height=\faHeight]{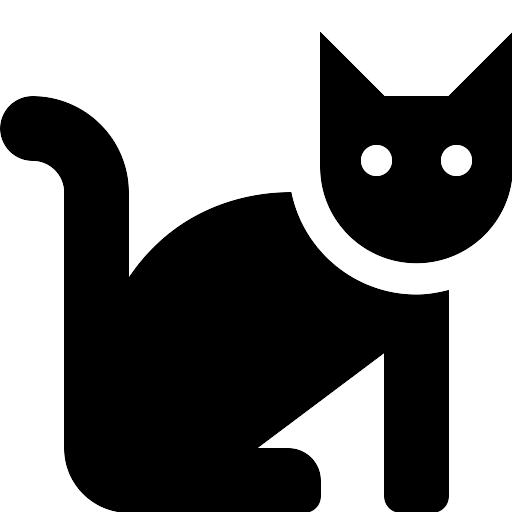}}
\newcommand{\faCow}{\includegraphics[height=\faHeight]{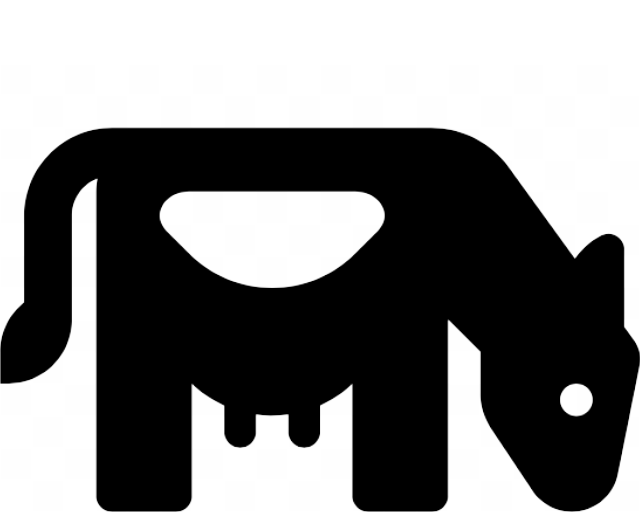}}
\newcommand{\faChair}{\includegraphics[height=\faHeight]{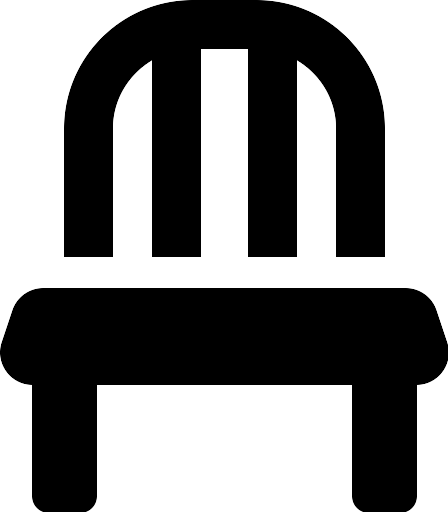}}
\newcommand{\faCouch}{\includegraphics[height=\faHeight]{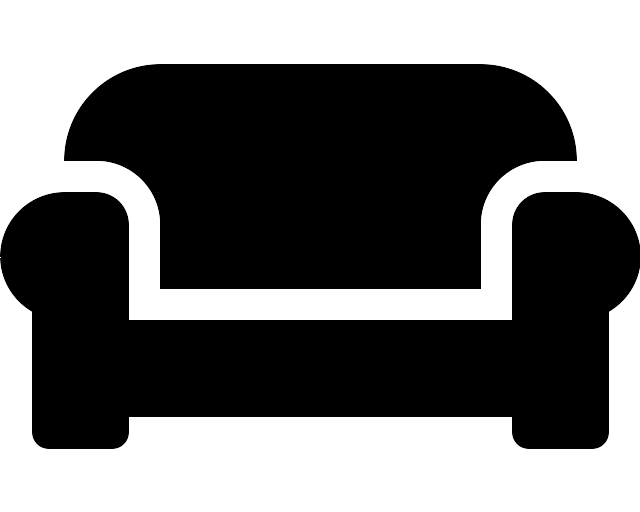}}

\newcommand{\faDog}{\includegraphics[height=\faHeight]{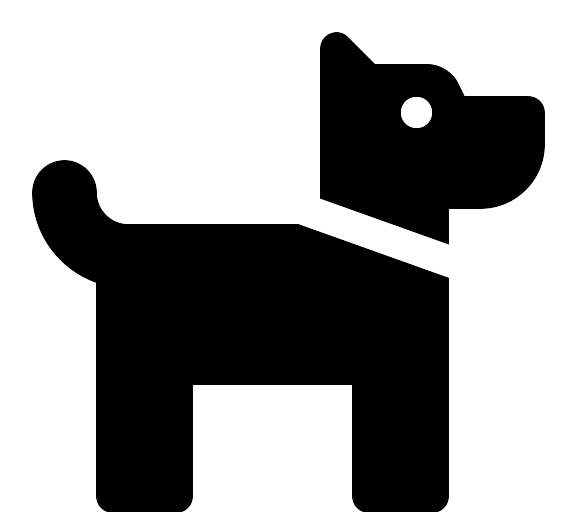}}

\newcommand{\faHorse}{\includegraphics[height=\faHeight]{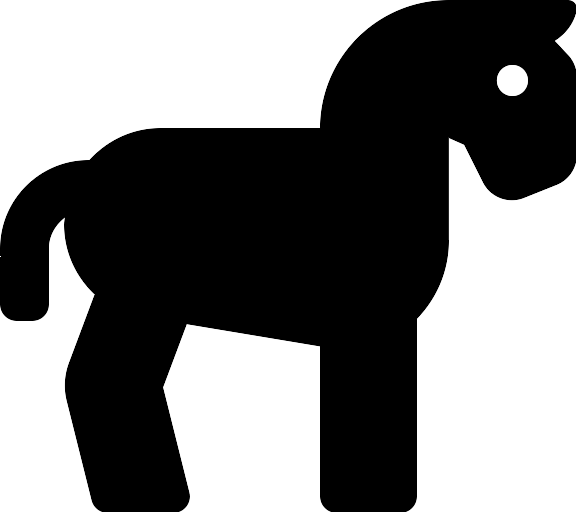}}

\newcommand{\faPlane}{\includegraphics[height=\faHeight]{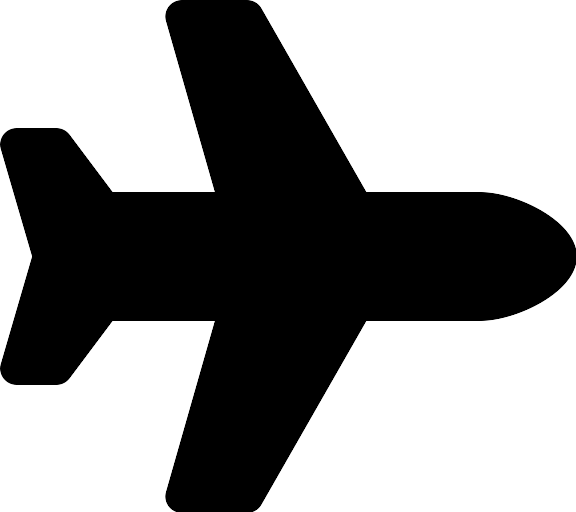}}
\newcommand{\faShip}{\includegraphics[height=\faHeight]{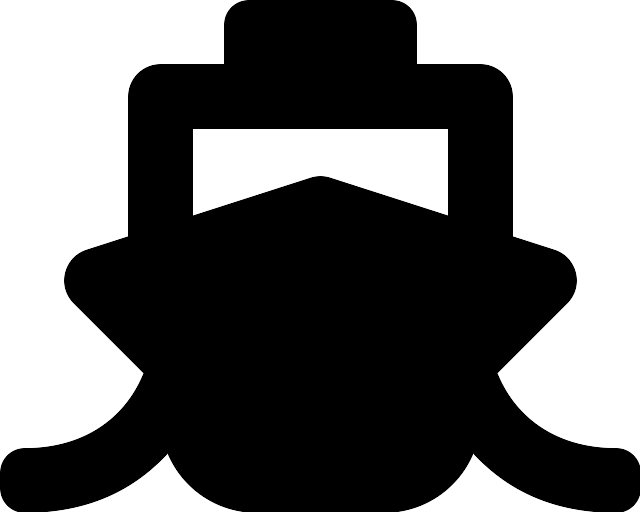}}
\newcommand{\faSheep}{\includegraphics[height=\faHeight]{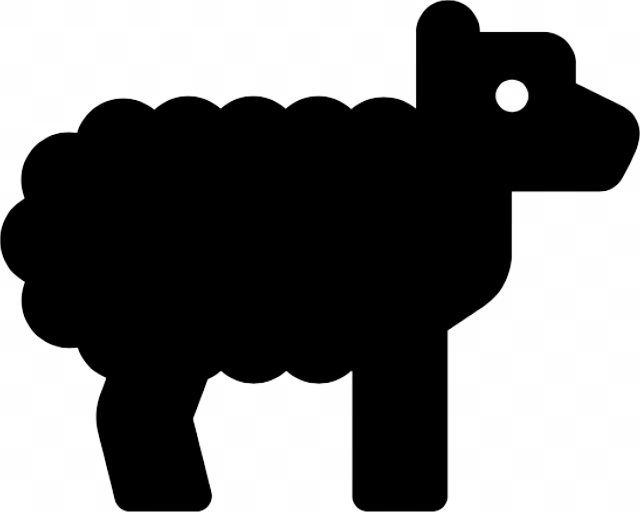}}

\newcommand{\faWalking}{\includegraphics[height=\faHeight]{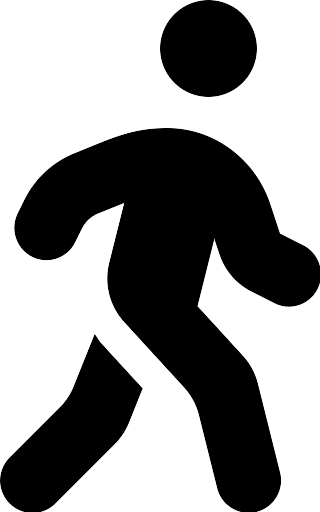}}

\usepackage[accsupp]{axessibility}  

\usepackage[pagebackref,breaklinks,colorlinks,citecolor=eccvblue]{hyperref}

\usepackage{orcidlink}

\begin{document}

\title{Dataset Enhancement with Instance-Level Augmentations} 

\titlerunning{InstanceAugmentation}

\author{Orest Kupyn \inst{1,2} \and
Christian Rupprecht \inst{1}}

\authorrunning{O.~Kupyn et al.}

\institute{Visual Geometry Group - University of Oxford \\
\email{\{okupyn, chrisr\}@robots.ox.ac.uk} \and
PiñataFarms AI \\}

\maketitle

\begin{center}
    \centering
    \captionsetup{type=figure}    
    \begin{subfigure}[t]{0.24\textwidth}
        \includegraphics[width=\textwidth]{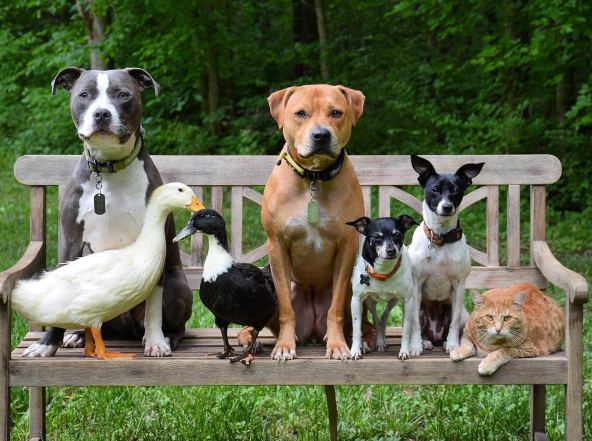}
    \end{subfigure}
    \begin{subfigure}[t]{0.24\textwidth}
        \includegraphics[width=\textwidth]{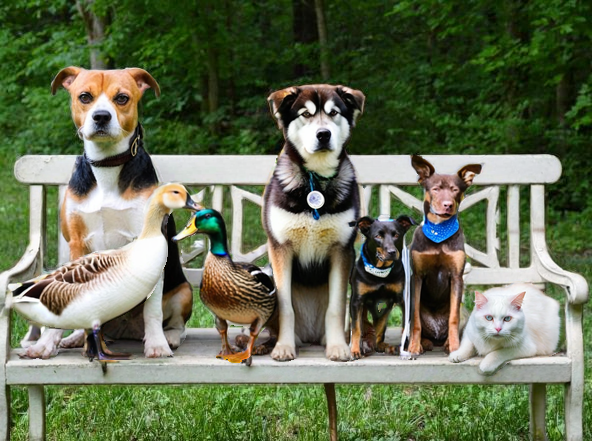}
    \end{subfigure}
    \begin{subfigure}[t]{0.24\textwidth}
        \includegraphics[width=\textwidth]{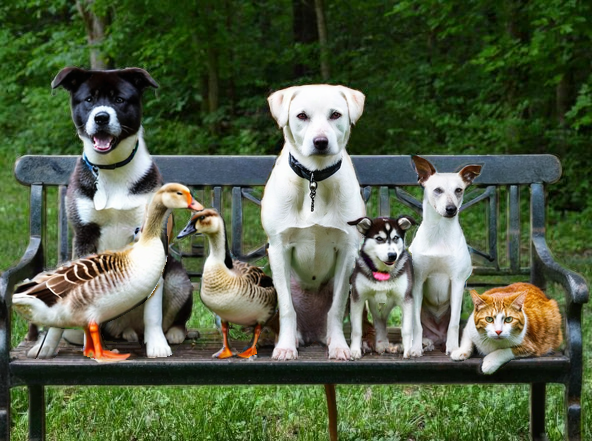}
    \end{subfigure}
    \begin{subfigure}[t]{0.24\textwidth}
        \includegraphics[width=\textwidth]{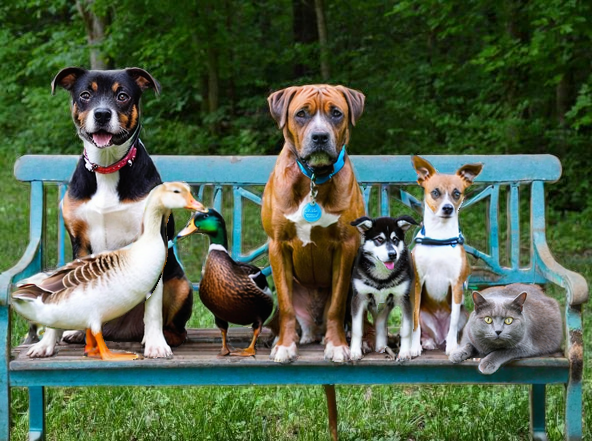}
    \end{subfigure}
    \begin{subfigure}[t]{0.24\textwidth}
        \includegraphics[width=\textwidth]{images/main_1_0.jpg}
    \end{subfigure}
    \begin{subfigure}[t]{0.24\textwidth}
        \includegraphics[width=\textwidth]{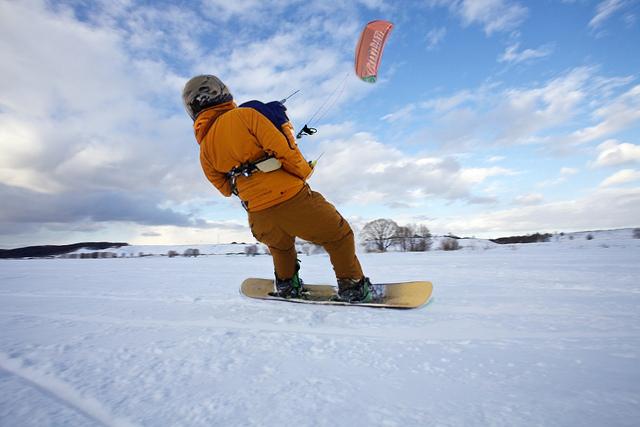}
    \end{subfigure}
    \begin{subfigure}[t]{0.24\textwidth}
        \includegraphics[width=\textwidth]{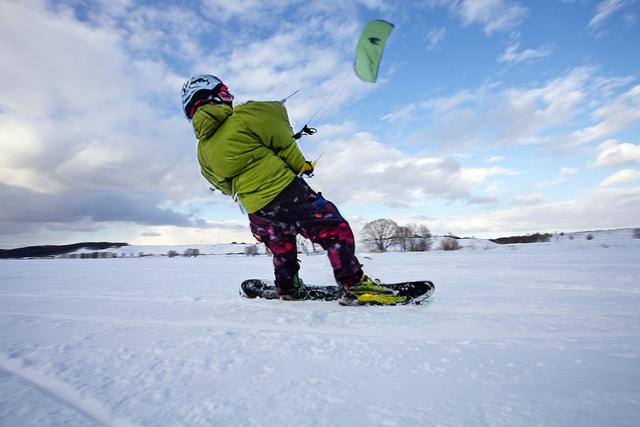}
    \end{subfigure}
    \begin{subfigure}[t]{0.24\textwidth}
        \includegraphics[width=\textwidth]{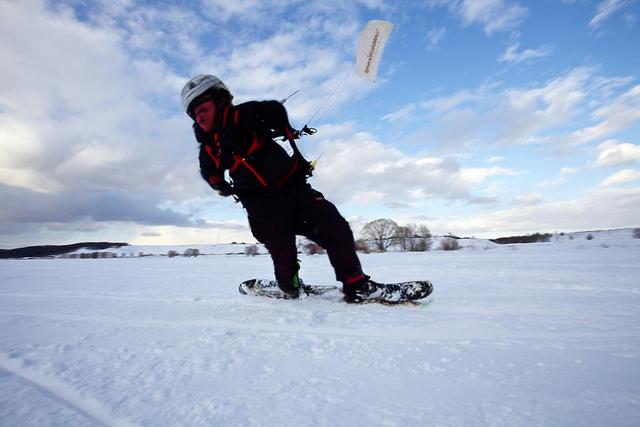}
    \end{subfigure}
    \caption{\textbf{Instance-Level Augmentations.} We augment images by redrawing individual objects in the scene retaining their original shape. This allows training with the unchanged class label (\eg class, segmentation, detection, \etc). The generations are highly diverse and match the scene composition. Guess the original in each row!}
\end{center}

\begin{abstract}
We present a method for expanding a dataset by incorporating knowledge from the wide distribution of pre-trained latent diffusion models.
Data augmentations typically incorporate inductive biases about the image formation process into the training (e.g. translation, scaling, colour changes, etc.).
Here, we go beyond simple pixel transformations and introduce the concept of instance-level data augmentation by repainting \textit{parts} of the image at the level of object instances. 
The method combines a conditional diffusion model with depth and edge maps control conditioning to seamlessly repaint individual objects inside the scene, being applicable to any segmentation or detection dataset.
Used as a data augmentation method, it improves the performance and generalization of the state-of-the-art salient object detection, semantic segmentation and object detection models.
By redrawing all privacy-sensitive instances (people, license plates, etc.), the method is also applicable for data anonymization.
We also release fully synthetic and anonymized expansions for popular datasets: COCO, Pascal VOC and DUTS. 
All datasets and the code is available at \href{https://github.com/KupynOrest/instance_augmentation}{instance augmentation}.
\end{abstract}
\vspace{-0.5cm}
\section{Introduction}
\label{sec:intro}
Deep learning has transformed computer vision research and applications.
Interestingly, methodological progress now advances hand in hand with the availability of large annotated datasets \cite{llm}.
This now shifts some of the focus from methods to their training data. 
Datasets have grown exponentially in size and complexity and are often just as important as the methods themselves.
From the one million images of ImageNet \cite{imagenet} to LAION-5B \cite{laion}, the number of samples has increased by three orders of magnitude. Large-scale datasets proved to be easily applicable to a wide range of computer vision transfer tasks \cite{radford2021learning, scalingviz, openclip}.
Yet, growing and curating such datasets at this scale is a cumbersome and costly challenge. 
Additionally, scraping large image collections from the internet does not adhere to ethical and data privacy standards. 
Moreover, in many fields, collecting large-scale datasets is impracticable due to the lack of available data and the complexity of the collection and annotation process (medical data, 3D, etc.).

In the wake of larger and larger datasets \cite{openimages, laion}, many older and/or smaller datasets have lost relevance purely due to their size even though they pose unique challenges, applications and benchmarks.
One of the main limiting factors of small datasets is their ability to train models that do not overfit.
Over the last decade, many ideas have been explored to overcome overfitting problems on small datasets.
Data augmentations such as colour shifts, flipping, crops, and rotations are being used to incorporate priors about geometric and lighting variations that we expect in the real world \cite{aug_survey}.
Their effectiveness is proven by their widespread use to date.
Image-level augmentations are geometric transformations or pixel manipulations, such as adding Gaussian blur or changing image contrast. On the level of objects, augmentations are also often simple: for example, copy-pasting objects between images \cite{copy_paste}.

Simple pixel manipulations only slightly expand a dataset's information and visual diversity.
Thus, several advanced techniques to mix images have been proposed \cite{cutmix, mixup, openmixup} with moderate success. Combining two or more images often creates visual artefacts that models can exploit to recognize artificial samples.
As a further improvement, generative models (trained on the to-be-augmented dataset) have been used to generate additional training samples \cite{datasetdm, handsoff}. 
However, as the generative model is trained on the same data, it cannot provide additional information beyond the dataset itself. 
Additionally, generating whole images also requires generating the accompanying annotations, which can be very challenging to do and verify.

To overcome the problem of annotation generation, we introduce a method that only regenerates a \textit{part} of an image at the level of object instances.
This allows us to retain the original data annotations (\eg class labels, segmentation, captions, etc.).
To significantly enhance the visual variability of the original data, we incorporate out-of-domain knowledge via a large diffusion-based generative model.
As these models have been trained on much larger volumes of data, their generations will naturally go beyond the variety originally contained in the dataset.
Another advantage of this approach is that it allows us to naturally combine real and generated data \textit{in the same image}. This exponentially increases the variations of samples that can be constructed from a single image. For example, an image with five objects has 32 variations of real/generated samples if we consider only one generation per object.
Finally, for privacy-sensitive instances such as people, license plates, etc. one can completely replace all real data in the dataset, strongly mitigating privacy concerns.

In this paper, we experiment with three tasks (object detection, semantic segmentation and saliency segmentation) and six datasets (MS-COCO \cite{mscoco}, PascalVOC \cite{voc}, DUTS \cite{duts}, ECSSD \cite{ecssd}, DUT-OMRON \cite{dut_omron} and HKU-IS \cite{hku_is}).
We show consistent improvements using our method to generate data for state-of-the-art methods for each task.
Further, we can show that fully replacing people in the training data with synthetic samples does not affect the final performance, which enables retrospectively improving privacy shortcomings of scraped internet datasets.
Together with the paper, we are releasing the code and generative variations of DUTS\cite{duts}, Pascal VOC \cite{voc} and COCO\cite{mscoco}.

\begin{figure*}[t!]
\includegraphics[width=0.99\textwidth]{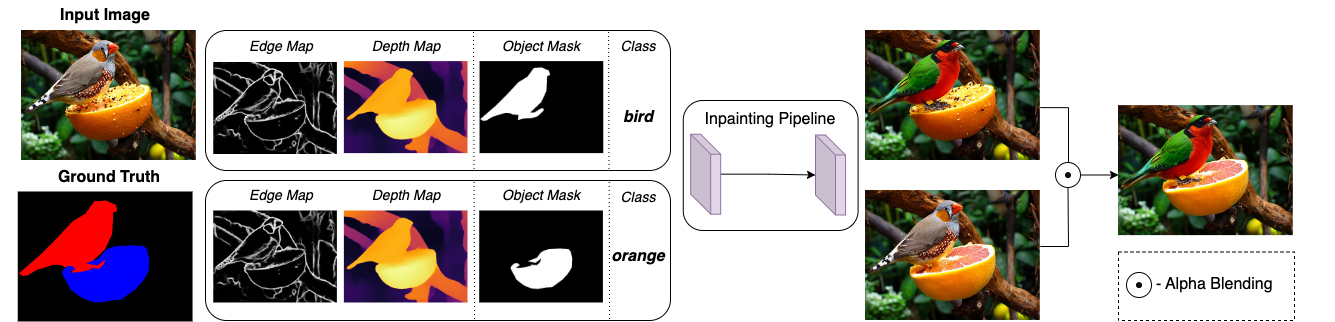}
\centering
\caption{\textbf{Overview.} Given an image and ground truth (or predicted) segmentation mask, we estimate depth and edge maps at the image level. The annotation is decomposed into the per-object binary masks and class, which together form the conditioning of the inpaining model. We redraw every instance and recombine them into a final image using alpha-blending sorted by depth. }
\label{fig:architecture}
\end{figure*}

\section{Related Work}
\label{sec:related_work}
This section provides an overview of related work on diffusion models, followed by a survey of the methods targeting synthetic data generation.

\paragraph{Image Generative Models.} 
Generative Adversarial Networks first introduced efficient sampling of high-resolution images with good perceptual quality \cite{gan, largegan, analyze_stylegan}. 
GAN models can generate visually plausible images but are challenging to optimize and struggle to capture the full data distribution \cite{unroll_gan}. Recently, diffusion probabilistic models \cite{ddpm, deepdm} have been introduced to match the underlying data distribution by learning to reverse a sample noising process. 
The high quality and stability of the training process quickly set diffusion models apart as a frontier in the field of image synthesis \cite{sdxl, hier_diff, cascadediff, text2img}. 
Text-to-image diffusion models condition the generation process by encoding text prompts into the latent vectors utilizing large pre-trained text encoders \cite{clip}. Latent diffusion models \cite{ldm} introduce a more general conditioning method fusing text embeddings, bounding boxes, or inpainting masks through cross-attention layers in a convolutional manner, achieving state-of-the-art results in image inpainting and class-conditional image synthesis performing diffusion steps in the latent image space \cite{latent_space}. ControlNet \cite{controlnet} and T2I-Adapter \cite{t2i} add spatial conditional control to diffusion models by encoding a target image representation in the form of edges, depth, segmentation, human pose, \etc.

\paragraph{Generative Models for Synthetic Datasets.} 
Early studies in synthetic data generation \cite{virtual, virtual_kitti} rely on 3D rendering engines to address common 2D vision problems. The process is limited by the domain of 3D models, cannot be generalized to complex real-world scenes and does require modifying the rendering engine for each specific dataset and subtask. For example, Virtual KITTI consists of exclusively street driving scenes for autonomous driving. 
In contrast, synthetic data generated using generation models using generative adversarial networks \cite{editgan, handsoff} are more flexible and generalize better to real-world images. Recently \cite{synth_saliency} combines diffusion embedding network with the BigGAN \cite{largegan} to generate synthetic images for the saliency detection. However, those methods primarily focus on learning the real dataset distribution, thus unable to incorporate new information into the dataset. Recently, \cite{imagenet_diversity, fake_imagenet, stablerep, synth_imagenet} improve the performance of classification models by generating synthetic data with latent diffusion models \cite{ldm}. However, the methods are limited to image classification.
Diffumask \cite{diffumask} and DatasetDM \cite{datasetdm} utilize large diffusion models to simultaneously generate synthetic images and annotations for semantic segmentation or depth prediction tasks. Conditioning on text embeddings from large language models \cite{clip} provides a mechanism to encode information outside of the original data distribution. Still, the methods do require finetuning on every specific dataset.
In contrast, our method directly employs pre-trained diffusion models without additional finetuning steps, not overfitting to the task-specific datasets and providing a way to efficiently mix real and synthetic annotations for training various vision models. It generalizes existing datasets by incorporating rich priors from the large-scale datasets used for image synthesis by augmenting image samples on an object level.

\section{Method}
\label{sec:method}

The goal of our method is to take an image $I \in \mathcal{I} = \mathbb{R}^{3 \times H \times W}$ with dimensions $H$ and $W$, and a set of annotations $\mathcal{Y}$, and generate a new image $I^* \in \mathbb{R}^{3 \times H' \times W'}$ with a mapping $F(I, \mathcal{Y}) = I^*$ preserving the structure of the scene and ground truth annotations.
This way, the sample $(I^*, \mathcal{Y})$ remains a valid training sample. The method should be able to generate high-fidelity image samples from a real-world distribution that are indistinguishable from the input image.
Since older datasets are often of low resolution, we allow for an increased image size $H' \ge H, W' \ge W$ for the generated image.

\subsection{Generation Pipeline}

To generate new samples $I^*$ we base the pipeline on a conditional latent diffusion model (LDM) \cite{ldm}.
Large-scale generative diffusion models have been trained on an extremely large variety of images, providing a strong signal to enhance the data distribution of any dataset.

Since a pure text-to-image generative model \cite{text2img} is not applicable to the task, we use an LDM trained for image inpainting to operate only on a region of interest.
For this purpose, we assume the following annotations to be available within $\mathcal{Y} = \{ (M_i, c_i) \}_{1 \leq i \leq N} $. For each of the $N$ objects in the image: a binary mask $M_i \in \mathcal{M} = \{0, 1\}^{H \times W}$ defining the segmentation of the object in the image, and a class label $c$ which can be free-form text. 
If this information is not already available in the dataset, it can be obtained by an off-the-shelf instance segmentation method. 

In general, the image inpainting process can be formulated as a function $G : \mathcal{I} \times \mathcal{M \times \mathcal{T}} \rightarrow \mathcal{I}$, that takes as input an image, a mask selecting the inpainting region and a text prompt that specifies what to draw. The output is an image with the masked region being redrawn.

A new image $I^*$ is generated by iteratively redrawing each object in the image.

\begin{equation}\label{eq:inpainting}
    I_i^* = G(I^*_{i-1}, M_i, T_i) .
\end{equation}

This iterative process starts with the original image $I_0^* = I$ and ends with every object being redrawn $I^*_N = I^*$. It requires a text-prompt $T_i$ for each object. In the simplest case the class name can be used as the text prompt $T_i = c_i$. We will explore more sophisticated choices in \cref{sec:method:prompting}.

In practice, two aspects are important to consider with the inpainting process. One is that the order in which the objects are being redrawn is arbitrary in \cref{eq:inpainting}, and does not take into account occlusions or overlapping objects, impacting the final result. The second is that repeatedly passing an image through an inpainting model significantly reduces the image quality. We will deal with both problems next.
\paragraph{Draw order.}
To ensure a reasonable blending of objects, we sort objects by their estimated relative distance to the camera. We use an off-the-shelf depth-estimation method DepthAnything \cite{depth_anything} to compute a (relative) depth map $D \in \mathbb{R}_+^{H \times W}$ and use the masks to compute a depth estimate $d_i = \sum_{u \in \Omega} M_i[u]D[u] \in \mathbb{R}_+$, where $\Omega = \{u,v | 1 \leq u \leq H, 1 \leq v \leq H \}$ is the set of all pixel locations in the image. 
We can then order the images back-to-front such that $d_i \ge d_{i+1}$ before inpainting.

\begin{figure}[htb]
\includegraphics[width=0.23\textwidth]{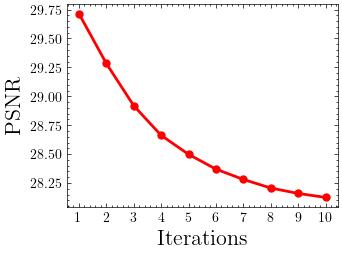}
\includegraphics[width=0.23\textwidth]{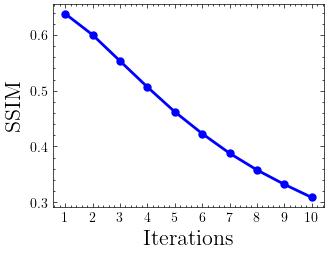}
\includegraphics[width=0.26\textwidth]{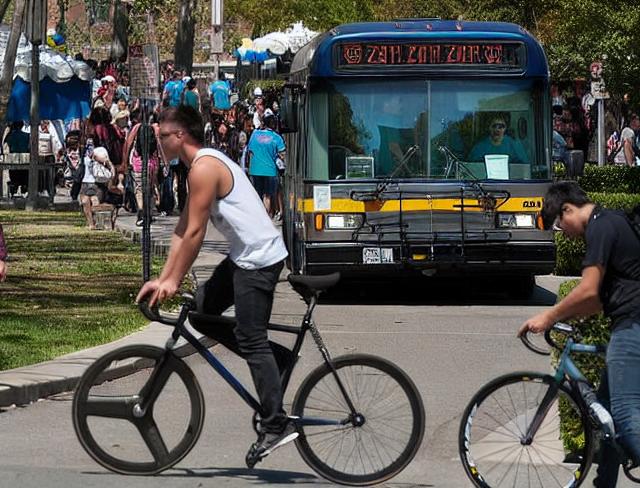}
\includegraphics[width=0.26\textwidth]{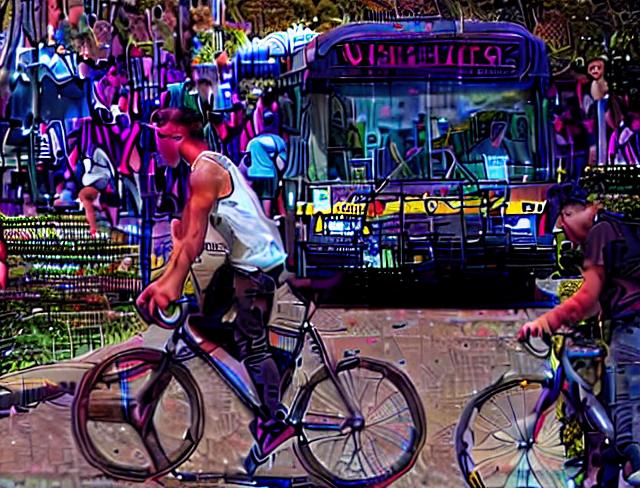}
\caption{\textbf{Noise Accumulation.} Images accumulate noise when repeatedly encoding and decoding to and from latent space. PSNR and SSIM compared to the original.}
\label{fig:vae_example}
\end{figure}

\paragraph{Iterative Redrawing.}
Current state-of-the-art inpainting models are not able to reconstruct the input image faithfully, even in regions that are not masked to be redrawn.
In LDMs this mainly comes from the fact that they operate in latent space instead of directly on images. 
Thus, images must be encoded and decoded from and to latent space. This process accumulates noise and leads to the image quality degrading for images with many objects.
We show this effect in \cref{fig:vae_example}, where we simply encode and decode to and from the latent space of the diffusion model without any noise sampling or diffusion steps. We measure the similarity of the decoded image after each step to the original image. 
Both PSNR and SSIM, decrease quickly with every iterative step.
We thus modify the process to always start with the original image:

\begin{equation}
    I_i^* = I^*_{i-1} \odot (\mathds{1} - M_i) + G(I, M_i, T_i) \odot M_i .
\end{equation}

This approach has two advantages. First, it does not require repeatedly applying the inpainter $G$ to the image, which would accumulate noise. On the other hand, this means that each object is drawn independently of the other objects, which allows generating an exponential number of $2^N$ variations of the image, where each object can be original or redrawn, with only $N$ applications of the inpainter.

\vspace{-0.3cm}
\vspace{-0.7cm}
\subsection{Better Inpainting} 
Naively employing an off-the-shelf inpainting model for our task can lead to several shortcomings. 
Using $G$ as a black box, there is no guarantee that the new object is of the same class as before, nor that it adheres to the original mask $M_i$.
We also observe that the LDM often tends to completely remove small objects, replacing them with background.

We thus make the following modifications to the inpainter $G$.
Several extensions to LDM provide more fine-grained control of the generated output. Specifically, ControlNet \cite{controlnet} injects additional conditions into the decoder blocks of the LDM generator.
We can thus condition the generation not only on the mask $M_i$, but also on the already computed depth map $D$ and an edge map $E$ that significantly decreases the potential discrepancy between the generated and original object contours. We compute the edge map using HED~\cite{hed}.

Interestingly, \cite{controlnet} has been trained on the non-inpainting version of \cite{ldm}, since the inpainting model was finetuned from the same baseline text to image model, they can still be used together \cite{finetune}.  

\subsection{Prompt Engineering}\label{sec:method:prompting}
While we can simply condition the inpainting model on the class information ($T_i = c_i$), several improvements can be made to both increase the quality as well as the control for a diversity of the generated objects.

The diffusion model provides an efficient way to add additional conditioning to the generation by modifying the input text prompt. Trained on the billions of image-text description pairs from the internet, using CLIP \cite{clip} as a text encoder, the model can encode the pose, shape, visual appearance, colour, and lightning of the object simultaneously. Our goal is to generate highly diverse samples. We exploit the useful properties of the text encoder to improve the visual quality and the diversity of the generated samples.

\paragraph{Object Description.} Following \cite{fake_imagenet}, we extend the label $c_i$ with its description. Each class in datasets such as COCO or ImageNet is associated with one or more synsets, \ie entities, in the WordNet \cite{wordnet} graph. We use the synset lemmas corresponding to each class to extend the class name.
This additional description of the object provides a more precise textual representation of the object in CLIP space. Empirically, we observe that it stabilizes the performance of the diffusion model and prevents the cases where an incorrect or no object at all is generated due to weak text embedding. 

\paragraph{Color and Lighting.} 
We manually select categories that tend to appear in many colour variations in the real world, such as cars, backpacks, snowboards, \etc.
For every such object, we randomly sample from a list of colours (\eg, blue, red, green, pink, \etc) and qualifiers (dark, light, natural, \etc) and add it as part of the prompt 
Additionally, we randomly select and add a lighting condition to the prompt (\eg, sunlight, dramatic lighting, soft lighting, \textit{none}, \etc).
We extensively experiment with these choices in \cref{sec:exp:prompting} and in the supplement.
Overall, our prompt variations increase the diversity across all classes.

\paragraph{Generating People.} 
``Person'' is the most visually complex and diverse class in common computer vision benchmarks with 15.5\% of DUTS and 30.5\% of COCO containing people in different scenes. We observe that for many cases of complex real-world scenes, the above depth and edge maps conditioning and simple prompting strategy are insufficient to generate a person that adheres to the scene's structure. 
Thus, to increase the descriptiveness of the prompt, for people, we use a visual question-answering model (BLIP-VQA \cite{blip}) to predict the action of the person and add it as a part of the prompt. 


\section{Datasets}

The method can generate images for an arbitrary dataset labelled with bounding boxes or segmentation masks, making it applicable for a wide range of tasks, including Object Detection, Semantic and Instance Segmentation, Panoptic Segmentation, \etc. 
For our experiments, we generate synthetic versions for three popular datasets: DUTS \cite{duts}, Pascal VOC \cite{voc} and MS-COCO \cite{mscoco}.
We run our generation pipeline up to three times for each object to obtain three synthetic replacements of the original instance.
This drastically increases the set of potential variations depending on the number of objects per image.
Naturally, we only generate novel data for the training set and leave the validation and test set untouched not to affect benchmark results.

For additional safety, we keep regenerating objects that do not pass an NSFW safety filter \cite{nsfw} with a strict threshold minimising the false negative rate.
The filter also detects NSFW content in the original datasets (which should not have been included in the dataset in the first place). 
Our redrawing method rectifies this. We will release all datasets and the code to generate them.

\paragraph{DUTS.} 
\cite{duts} is the largest benchmark dataset for salient object detection. 
Yet it still only contains 10,553 training and 5,019 test images of various real-world scenes and types of objects.
The annotations consist of a single saliency map per image and do not contain class labels or names. 
To construct the text prompt for the pipeline, we crop the image by the bounding rectangle of the binarized saliency map and use the BLIP-VQA \cite{blip} model to predict the object name in an open vocabulary setting. 
The dataset is of relatively low resolution, which we also improve with our method.

Since salient object segmentation highly depends on predicting accurate object boundaries, we add an optional mask refinement stage to preserve the sharpness and high quality of the masks.
To this end, we crop every generated object from the images in the train set using its corresponding bounding rectangle of the saliency map. 
We use TRACER-7 \cite{tracer} as an off-the-shelf segmentation model to obtain high-resolution, tight object crop saliency maps. 
This removes the potential mismatch between the generated object and its original annotations. 

\vspace{-0.3cm}
\begin{figure}[htb]
\begin{subfigure}[t]{0.32\textwidth}
 \includegraphics[width=\textwidth]{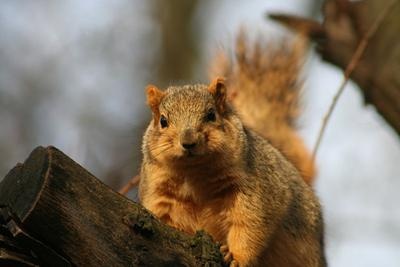}
 \end{subfigure}
\begin{subfigure}[t]{0.32\textwidth}
 \includegraphics[width=\textwidth]{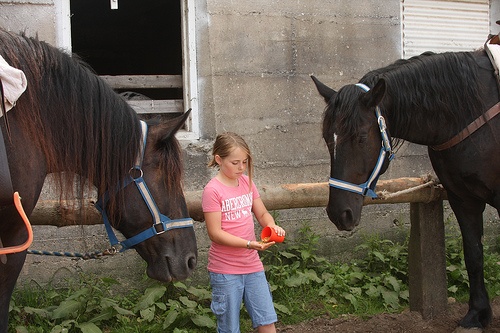}
 \end{subfigure}
\begin{subfigure}[t]{0.32\textwidth}
 \includegraphics[width=\textwidth]{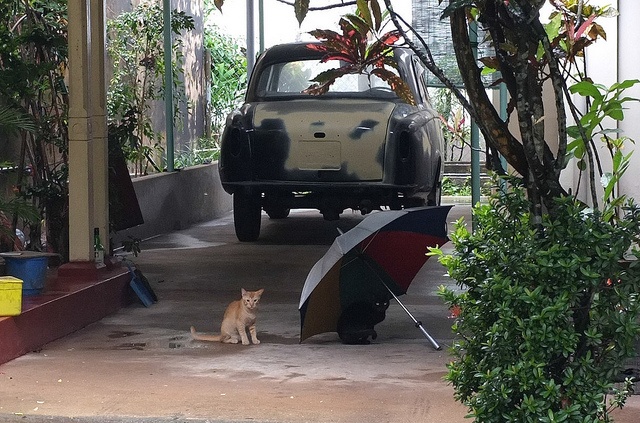}
 \end{subfigure}
 \begin{subfigure}[t]{0.32\textwidth}
 \includegraphics[width=\textwidth]{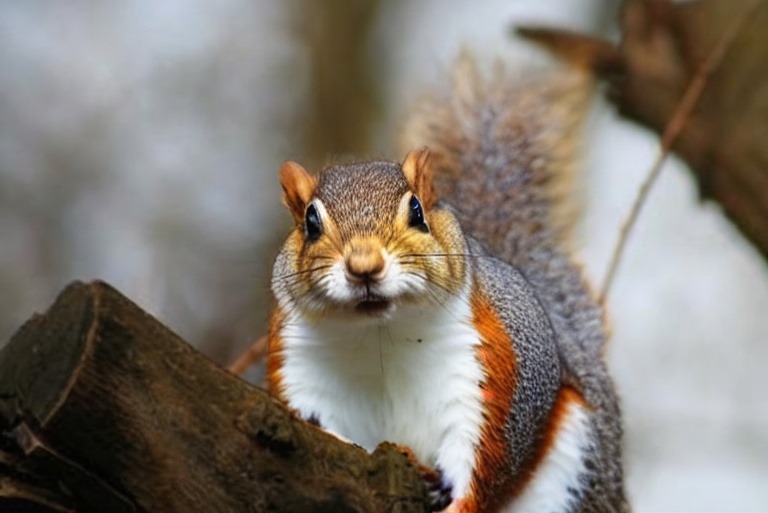}
 \caption{DUTS}
 \end{subfigure}
\begin{subfigure}[t]{0.32\textwidth}
 \includegraphics[width=\textwidth]{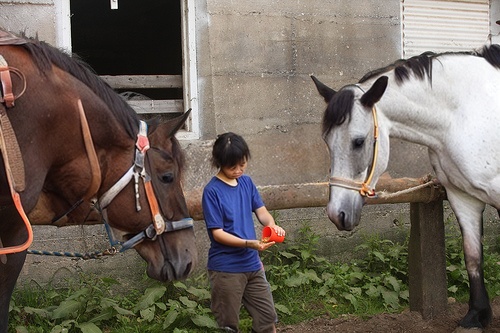}
 \caption{VOC}
 \end{subfigure}
\begin{subfigure}[t]{0.32\textwidth}
 \includegraphics[width=\textwidth]{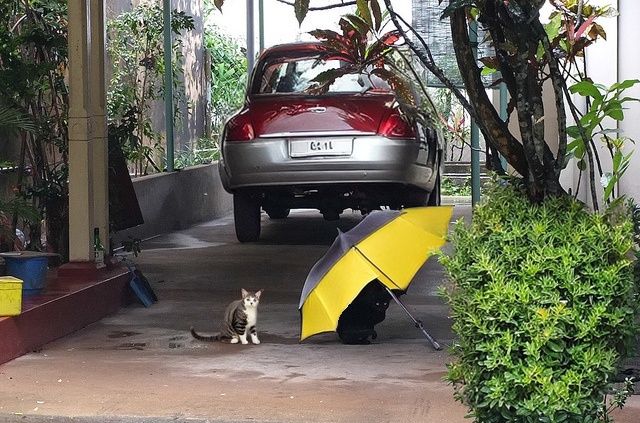}
 \caption{COCO}
 \end{subfigure}
\centering
\caption{\textbf{Qualitative Evaluation.} The examples of method performance on three different datasets. The method generalize well to the complex scenes and datasets with no ground truth instance labels.}
\label{fig:examples}
\end{figure}
\vspace{-0.5cm}
\paragraph{MS-COCO.} 
The MS-COCO dataset \cite{mscoco} is a standard benchmark for evaluating object detection, instance segmentation, and other tasks. It includes complex everyday scenes containing common objects in their natural context. 


\begin{figure}[htb]
\begin{subfigure}[t]{0.23\textwidth}
 \includegraphics[width=\textwidth]{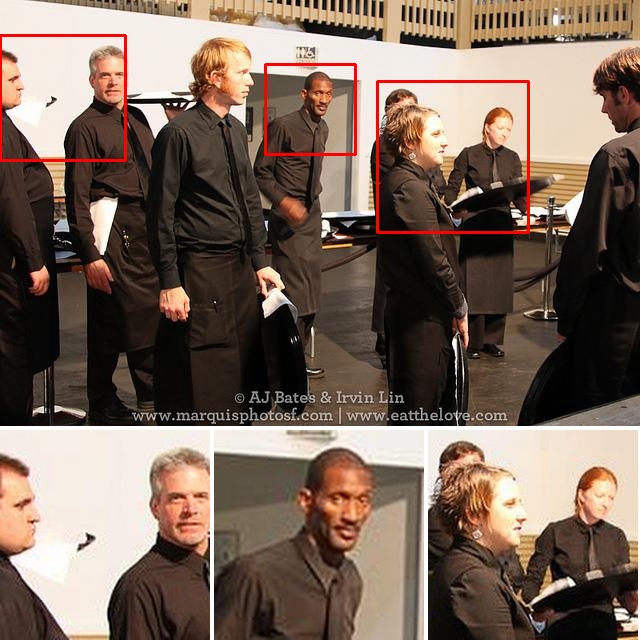}
 \caption{Original}
 \end{subfigure}
\begin{subfigure}[t]{0.23\textwidth}
 \includegraphics[width=\textwidth]{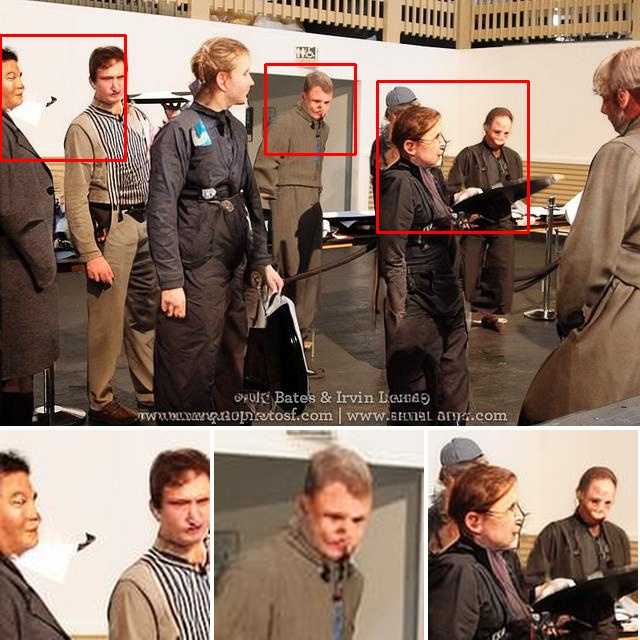}
 \caption{Generated}
 \end{subfigure}
\begin{subfigure}[t]{0.23\textwidth}
 \includegraphics[width=\textwidth]{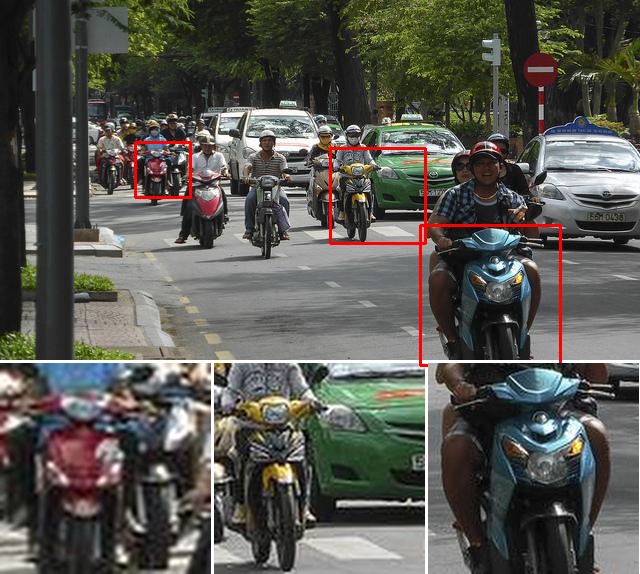}
 \caption{Original}
 \end{subfigure}
\begin{subfigure}[t]{0.23\textwidth}
 \includegraphics[width=\textwidth]{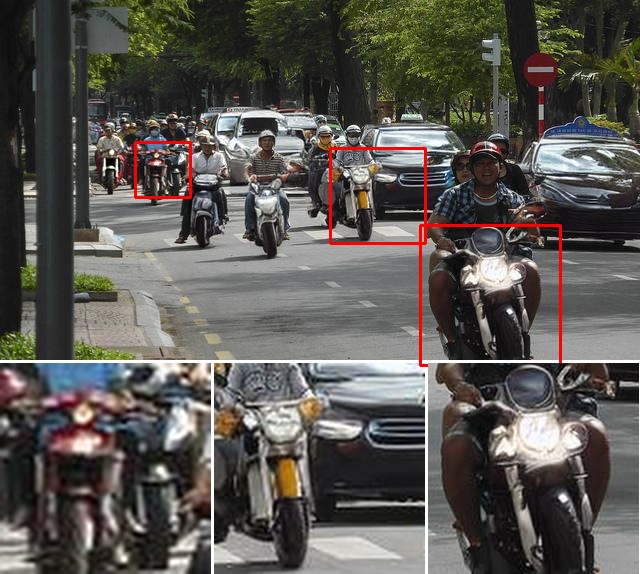}
 \caption{Generated}
 \end{subfigure}
\centering
\caption{\textbf{Data Anonymization.} The method efficiently repaints each annotation in the complex scenes, strongly mitigating privacy concerns for sensitive instances such as people or cars.}
\label{fig:anonymization}
\end{figure}

\paragraph{Pascal VOC.} 
The PASCAL Visual Object Classes (VOC) dataset \cite{voc} contains images 
with pixel-level segmentation annotations, bounding box annotations, and object class annotations. This dataset has been widely used as a benchmark for object detection, semantic segmentation, and classification tasks.

\paragraph{Dataset Anonymization.}
In addition to the generic version of the datasets we generate, we use our method to improve the privacy aspect of both datasets. 
For DUTS, we manually select all images containing people in the dataset and repaint them with ``virtual'' people. We thus generate an anonymized version of the dataset, which can be used by itself with all original images and annotations (except people) or in conjunction with the generated version.

For the COCO dataset, we generate two additional versions: anonymizing people and cars (for personal information such as license plates). 
For this task, to ensure every object (and not only the ten largest ones) in the dataset is anonymized, we process every object, including small instances. For objects with an area lower than $32*32$px the image is first cropped and upsampled to 512px resolution to match the minimum resolution the generator model is trained on. See \cref{fig:anonymization} for visual examples of repainting people and vehicles.

\section{Experimental Evaluation}
\label{sec:exp_eval}

We evaluate the efficiency of our method (i) as a data augmentation technique, (ii) as a pipeline for data anonymization, and (iii) its impact on the model generalization. To address (i) we train the state-of-the-art saliency segmentation, semantic segmentation and object detection models on the combined original and generated by our methods versions of DUTS, VOC and COCO datasets, respectively, and report the findings in the \ref{t:aug}. To test the effectiveness of data anonymization (ii) we train the object detection models on the generated version of COCO datasets with all cars and people entirely replaced with the synthetic objects and compare it with the original model \ref{t:anonym}. To test generalization (iii) we analyze the performance of the models trained on combined datasets on the established benchmarks for salient object detection, detailed in \ref{t:saliency}.

\subsection{Data Augmentation}

\paragraph{Object Detection.} The effect of the instance augmentations is tested on a large-scale complex COCO dataset. We evaluate the performance on three different object detection architectures, transformer-based: Deformable-DETR (single scale) \cite{deformabledetr}, RT-DETR (with ResNet-50) backbone \cite{rtdetr} and an anchor-free YOLOv5m \cite{yolov5}. On each training iteration, every instance in an image is repainted with 30\% probability. This ensures a high diversity of training samples and reduces the overfitting. The results (\cref{t:coco_main}) show that adding augmented data consistently improves the performance of object detection models. 

Additionally, we evaluate the effectiveness of augmented data in a data-spare setting, with only a subset of the training data available. We randomly select subsample 10\%, 25\%, 50\%, 75\% and 100\% of the COCO training set, covering all categories. \Cref{t:coco_20} shows the consistent boost of \textit{+2-3 AP} for each setting.

\begin{table}[t]
\centering
\begin{minipage}{\textwidth}
    \begin{minipage}{0.50\textwidth}
\footnotesize
\setlength{\tabcolsep}{1.5pt}
\centering
\begin{tabular}{lc | ccc}
\toprule
 \textbf{Model} & \textbf{Data} & \textbf{AP} & \textbf{AP$_{50}$} & \textbf{AP$_{75}$}\\
\midrule
Def.-DETR & orig. & 39.3 & 60.0 & 42.0\\
Def.-DETR & ours & \textbf{40.5} & \textbf{60.2} & \textbf{43.4}\\
\midrule
RT-DETR & orig. & 51.4 & 69.6 & 55.4\\
RT-DETR & ours & \textbf{52.4} & \textbf{69.7} & \textbf{56.5}\\
\midrule
YOLOv5m & orig. & 44.1 & 63.4 & 47.8\\
YOLOv5m & ours & \textbf{45.7} & \textbf{64.0} & \textbf{49.7}\\
\bottomrule
\end{tabular}
\captionof{table}{ \textbf{Object Detection}. Our instance augmentations improve all object detectors on MS-COCO \cite{mscoco}.}
\label{t:coco_main}
    \end{minipage} \hfill
    \begin{minipage}{0.45\textwidth}
\footnotesize
\setlength{\tabcolsep}{1.5pt}
\centering
\begin{tabular}{l ccccc}
\toprule
Data & \textbf{10\%} & \textbf{25\%} & \textbf{50\%} & \textbf{75\%} & \textbf{100\%}\\
\midrule
orig. & 25.8 & 34.2 & 39.1 & 41.2 & 44.1\\
ours & \textbf{27.9} & \textbf{36.1} & \textbf{40.3} & \textbf{42.5} & \textbf{45.7}\\
\bottomrule
\end{tabular}
\captionof{table}{ \textbf{Data-Sparse Object Detection.} Our data improves the performance of YOLOv5 detector consistently, even with limited amount of training data available.}
\label{t:coco_20}
    \end{minipage}
\end{minipage}
\end{table}

\paragraph{Semantic Segmentation.} To demonstrate the impact of augmentations when no ground truth instance masks are available we evaluate the performance of semantic segmentation models on Pascal VOC augmented set. We utilize the pseudo-ground-truth instances from \cite{irn}. To generate the augmented images. Following \cite{diffumask} we first train Mask2Former \cite{mask2former} with ResNet-50 backbone on the synthetic dataset and then finetune the model on the subset of original VOC data. \ref{t:voc} show significant improvement of \textbf{15.5 mIoU} with even larger improvements on challenging categories (\textbf{42.4} vs \textbf{14.7} mIoU for chair). The main limitation of generating fully synthetic samples is what motivates our approach: diffusion models do not perform well in generating complex scenes with multiple foreground/background objects, even with additional guidance. Further, the model finetuned on the original data achieves superior performance on almost all categories comparing both to the baseline and the Diffumask method.
\begin{table*}[h!]
\tiny
\footnotesize
\centering
\setlength{\tabcolsep}{1pt}
\renewcommand{\arraystretch}{1.2}
\resizebox{\linewidth}{!}{\begin{tabular}{cc|ccccccccccccc|c}
\textbf{Real} &\textbf{Syn.} & \faPlane & \faBird & \faShip & \faBus & \faCar & \faCat &	\faChair & \faCow & \faDog & \faHorse & \faWalking & \faSheep & \faCouch & \textbf{mIoU} \\
\toprule
 VOC & \xmark & 87.5 & \textbf{94.4} & 70.6 & \textbf{95.5} & 87.7 & 92.2 & \textbf{44.0} & 85.4 & 89.1 & 82.1 & \textbf{89.2} & 80.6 & 53.6 & 77.3 \\
\midrule
\xmark & \cite{diffumask} & 80.7 & 86.7 & 56.9 & 81.2 & 74.2 & 79.3 & 14.7 & 63.4 & 65.1 & 64.6 & 71.0 & 64.7 & 27.8 & 57.4 \\
\xmark & Ours & 86.5 & 89.1 & 71.7 & 85.9 & 80.7 & 92.5 & 42.2 & 66.0 & 87.2 & 74.8 & 86.3 & 72.0 & 43.5 & 72.9 \\
\midrule
VOC & \cite{diffumask} & 85.4 & 92.8 & \textbf{74.1} & 92.9 & 83.7 & 91.7 & 38.4 & \textbf{86.5} & 86.2 & 82.5 & 87.5 & 81.2 & 39.8 & 77.6 \\
VOC & Ours & \textbf{89.2} & 89.4 & 69.0 & 92.3 & \textbf{87.8} & \textbf{93.6} & 40.5 & 79.8 & \textbf{89.6} & \textbf{86.7} & 88.9 & \textbf{85.2} & \textbf{61.5} & \textbf{78.2} \\
\bottomrule
\noalign{\smallskip}
\end{tabular}}
\caption{ \textbf{Semantic Segmentation Evaluation.} We evaluate the performance of Mask2Former \cite{mask2former} with ResNet-50 backbone on VOC \cite{voc} dataset. Training on augmented data improve the results by a wide margin comparing to \cite{diffumask}, while finetuning on a part of original dataset further improves model performance.}
\label{t:voc}
\vspace{-0.8cm}
\end{table*}
\paragraph{Salient Object Detection.} The performance of the salient object detection models is evaluated on four popular datasets, namely ECSSD \cite{ecssd} with 1000 images of relatively complex backgrounds, DUT-OMRON \cite{dut_omron} with 5168 images that include one or more salient objects with rather complex backgrounds, HKU-IS \cite{hku_is} with 4,447 images, that include two or more objects with various backgrounds and DUTS \cite{duts} with 15,572 images which is the largest available dataset for training divided into 10,553 training images (DUTS-TR) and 5,019 testing images (DUTS-TE). All datasets are labelled with pixel-wise ground truth.


We use three saliency segmentation models: U2Net \cite{u2net}, F3-Net \cite{f3net} and TRACER \cite{tracer}. 
Specifically, we choose the U2Net-lite version as a simpler architecture and the more complex multi-head TRACER model with an EfficientNet-4 \cite{effnet} backbone to show the impact of the data augmentation on different model sizes and architectures. 


\begin{table*}[t]
\footnotesize
\centering
\setlength{\tabcolsep}{0.5pt}
\resizebox{\linewidth}{!}{\begin{tabular}{lc|ccc|ccc|ccc|ccc}
\toprule
 & & \multicolumn{3}{c}{ECCSD} & \multicolumn{3}{c}{DUTS-TE} & \multicolumn{3}{c}{DUT-OMRON} & \multicolumn{3}{c}{HKU-IS}\\
 \textbf{Model} & \textbf{Data} & \textbf{$\boldsymbol{F_{max}}$}{$\uparrow$} & {\textbf{MAE}{$\downarrow$}} & $\boldsymbol{S_m}${$\uparrow$} & \textbf{$\boldsymbol{F_{max}}$}{$\uparrow$} & {\textbf{MAE}{$\downarrow$}} & $\boldsymbol{S_m}${$\uparrow$} & \textbf{$\boldsymbol{F_{max}}$}{$\uparrow$} & {\textbf{MAE}{$\downarrow$}} & $\boldsymbol{S_m}${$\uparrow$} & \textbf{$\boldsymbol{F_{max}}$}{$\uparrow$} & {\textbf{MAE}{$\downarrow$}} & $\boldsymbol{S_m}${$\uparrow$}\\
\midrule
U2Net & orig. & 0.944 & 0.052 & 0.900 & 0.863 & 0.066 & 0.836 & 0.835 & 0.075 & 0.819 & 0.930 & 0.043 & 0.895\\
U2Net & ours & \textbf{0.948} & \textbf{0.047} & \textbf{0.908} & \textbf{0.874} & \textbf{0.061} & \textbf{0.844} & \textbf{0.848} & \textbf{0.069} & \textbf{0.827} & \textbf{0.935} & \textbf{0.040} & \textbf{0.900} \\
\hline
F3-Net & orig. & 0.955 & 0.035 & 0.924 & 0.899 & 0.038 & 0.887 & 0.831 & \textbf{0.054} & 0.835 & 0.942 & \textbf{0.029} & 0.918\\
F3-Net & ours & \textbf{0.962} & \textbf{0.033} & \textbf{0.929} & \textbf{0.907} & \textbf{0.037} & \textbf{0.890} & \textbf{0.850} & 0.055 & \textbf{0.843} & \textbf{0.947} & \textbf{0.029} & \textbf{0.922} \\
\hline
TRACER-4 & orig. & 0.956 & 0.027 & 0.929 & 0.911 & 0.029 & 0.896 & 0.847 & 0.048 & 0.848 & 0.944 & 0.024 & 0.921\\
TRACER-4 & ours & \textbf{0.960} & \textbf{0.026} & \textbf{0.933} & \textbf{0.918} & \textbf{0.026} & \textbf{0.905} & \textbf{0.848} & \textbf{0.045} & \textbf{0.853} & \textbf{0.948} & \textbf{0.022} & \textbf{0.928} \\
\bottomrule
\noalign{\smallskip}
\end{tabular}}
\caption{ \textbf{Salient Object Detection Evaluation.} Comparison of performance with five existing methods on five benchmark datasets. The best results per method pair are highlighted. Our method improves the performance in 34/36 metrics.}
\label{t:saliency}
\vspace{-0.6cm}
\end{table*}

\begin{table*}[t]
\footnotesize
\centering
\setlength{\tabcolsep}{0.5pt}
\resizebox{\linewidth}{!}{\begin{tabular}{lccc|ccc|ccc|ccc|ccc}
\toprule
 & & & & \multicolumn{3}{c}{ECCSD} & \multicolumn{3}{c}{DUTS-TE} & \multicolumn{3}{c}{DUT-OMRON} & \multicolumn{3}{c}{HKU-IS}\\
 \textbf{Model} & \textbf{Aug} & \textbf{Real} & \textbf{Syn.} & \textbf{$\boldsymbol{F_{max}}$}{$\uparrow$} & {\textbf{MAE}{$\downarrow$}} & $\boldsymbol{S_m}${$\uparrow$} & \textbf{$\boldsymbol{F_{max}}$}{$\uparrow$} & {\textbf{MAE}{$\downarrow$}} & $\boldsymbol{S_m}${$\uparrow$} & \textbf{$\boldsymbol{F_{max}}$}{$\uparrow$} & {\textbf{MAE}{$\downarrow$}} & $\boldsymbol{S_m}${$\uparrow$} & \textbf{$\boldsymbol{F_{max}}$}{$\uparrow$} & {\textbf{MAE}{$\downarrow$}} & $\boldsymbol{S_m}${$\uparrow$}\\
\midrule
F3-Net & \xmark & \xmark & \cite{synth_saliency} & 0.938 & 0.039 & 0.913 & 0.856 & 0.053 & 0.851 & 0.793 & 0.077 & 0.802 & 0.932 & 0.032 & 0.909\\
F3-Net  & \xmark & \xmark & ours & \textbf{0.959} & \textbf{0.036} & \textbf{0.923} & \textbf{0.903} & \textbf{0.040} & \textbf{0.880} & \textbf{0.841} & \textbf{0.058} & \textbf{0.829} & \textbf{0.949} & \textbf{0.029} & \textbf{0.919}\\
\hline
TRACER-0 & \xmark & \cmark & \xmark & 0.943 & 0.035 & 0.912 & 0.884 & 0.035 & 0.874 & 0.816 & 0.050 & 0.825 & 0.929 & 0.031 & 0.904 \\
TRACER-0 & \cmark & \cmark & \xmark & \textbf{0.949} & 0.032 & 0.919 & 0.889 & 0.035 & 0.880 & \textbf{0.833} & 0.051 & 0.836 & 0.933 & 0.028 & 0.912\\
TRACER-0 & \xmark & \cmark & ours & \textbf{0.949} & 0.033 & 0.919 & 0.891 & 0.034 & 0.879 & 0.826 & \textbf{0.048} & 0.834 & 0.932 & 0.029 & 0.909\\
TRACER-0 & \cmark & \cmark & ours & \textbf{0.949} & \textbf{0.031} & \textbf{0.923} & \textbf{0.892} & \textbf{0.033} & \textbf{0.885} & 0.829 & 0.049 & \textbf{0.839} & \textbf{0.937} & \textbf{0.027} & \textbf{0.916}\\
\bottomrule
\end{tabular}}
\caption{ \textbf{Data Augmentation.} Comparison of the influence of standard augmentations (Aug) and the influence of including real samples (Real) as well as a comparison with the augmentation strategy in \cite{synth_saliency}.}
\label{t:aug}
\vspace{-0.8cm}
\end{table*}

\Cref{t:saliency} shows saliency segmentation results of the three models on the original and our version of each dataset. Using our data pipeline over the original images improves the performance in almost all metrics across five datasets and three models with up to 8\% in error reduction.
We note that all models are trained on DUTS (either the original data or ours). Thus, this benchmark also measures generalization, which is one of the strengths of our method.

We also investigate how traditional data augmentations (including flips, rotations, adding noise, blur, brightness, contrast, \etc) as implemented in \cite{albu} affect the results.
In \cref{t:aug}, we analyse the influence of using augmentations, real data and or synthetic data.
Additionally, we compare our synthetic data generation pipeline to the synthetic dataset of \cite{synth_saliency}.
Classical image augmentations are complementary to our method and can be used in conjunction. Further, our synthetic data performs better than \cite{synth_saliency}, especially for generalization.

\begin{table*}[t]
\footnotesize
\centering
\resizebox{\linewidth}{!}{\begin{tabular}{lc|ccc|ccc|ccc}
\toprule
& & \multicolumn{3}{c}{all classes} & \multicolumn{3}{c}{people only} & \multicolumn{3}{c}{cars only}\\
Model & Dataset & AP & AP$_{50}$ & AP$_{75}$ & AP & AP$_{50}$ & AP$_{75}$ & AP & AP$_{50}$ & AP$_{75}$\\
\midrule
Deformable-DETR & Real & 39.3 & 60.0 & 42.0 & 49.7 & 78.5 & 52.7 & 39.5 & 68.5 & 39.0\\
Deformable-DETR & Anonymized People & 38.3 & 58.7 & 40.6 & 45.6 & 75.2 & 47.1 & - & - & - \\
Deformable-DETR & Anonymized Cars & 38.7 & 59.5 & 41.5 & - & - & - & 37.2 & 65.6 & 36.7 \\
\bottomrule
\end{tabular}}
\caption{ \textbf{Data Anonymization.} Replacing all real data of people or cars only has a marginal impact on the overall performance. This is important as one might not even be interested in a detector for people, yet people are contained in the images. The replaced category slightly decreases in performance which is acceptable given the strong increase in privacy.}
\label{t:anonym}
\vspace{-0.6cm}
\end{table*}

\subsection{Data Anonymization}

The effectiveness of the introduced method for data anonymization is evaluated on COCO using our anonymized datasets, where all cars and people in the dataset were substituted with the synthetic versions. 
We train a Deformable-DETR \cite{deformabledetr} model for 50 epochs on the original MS-COCO \cite{mscoco} and the two anonymized variants and compute mean average precision and thresholded average precision metrics for all objects and people and cars separately. The model trained on purely synthetic subcategories without real people/car objects achieves comparable performance to the model trained on the original dataset within the category and on a complete validation set, proving the method can efficiently be employed to anonymize potentially sensitive data.
While the performance of the synthetic categories decreases slightly, the overall performance is almost unaffected (1 AP drop with 30.5\% of all instances repainted). As many applications are object-centric and do not depend on detecting humans accurately, the boost in privacy comes at no cost in performance.
Additionally, we measure the anonymization strength comparingly to state of the art face anonymization method \cite{ldfa} by calculating how often a replaced face can be matched with its original appearance using a face identification model. 
To this end, we use ArcFace \cite{arcface} on the pairs of original and generated images from the COCO dataset, validating whether people can be re-identified after applying our method. 
From 64115 images with people and 262465 faces, only 373 (0.14\%) of faces were re-identified vs. 2.8\% faces anonymized by LDFA \cite{ldfa}. This shows the importance of extending the method beyond face anonymization because it can replace all pixels covering a given individual. It is important as making only faces unidentifiable can be considered to be only pseudonymization, i.e., not a complete removal of personal information from an image.

Additionally, we test the effect on the DUTS dataset by anonymizing all people in  \cref{t:anonym_duts}. Similarly, the drop in performance is negligible, demonstrating high potential for tasks where preserving data privacy is crucial.

\begin{table}[t]
\footnotesize
\setlength{\tabcolsep}{1.5pt}
\centering
\begin{tabular}{lc cccc}
\toprule
 \textbf{Model} & \textbf{Dataset} & \textbf{$\boldsymbol{F_{max}}$}{$\uparrow$} & {\textbf{MAE}{$\downarrow$}} & $\boldsymbol{S_m}${$\uparrow$} & $\boldsymbol{F_{avg}}\uparrow$\\
\midrule
TRACER-0 & Real & 0.889 & 0.035 & 0.880 & 0.847\\
TRACER-0 & Anonymized & 0.886 & 0.036 & 0.876 & 0.846\\
\bottomrule
\end{tabular}
\caption{ \textbf{Anonymizing People in DUTS.} Training with only synthetic humans achieves comparable performance to the model trained on real images of humans.}
\label{t:anonym_duts}
\end{table}
\begin{figure}[t]
\captionsetup[subfigure]{labelformat=empty}
\begin{subfigure}[t]{0.24\textwidth}
 \includegraphics[width=\textwidth]{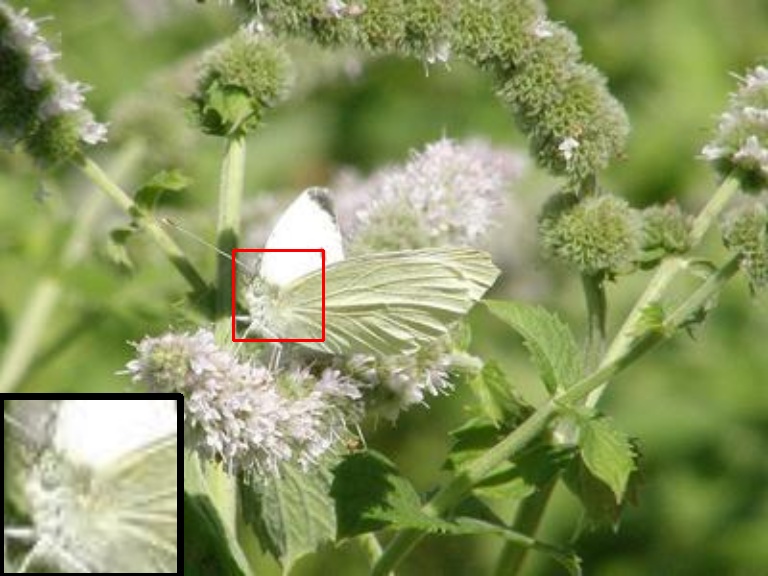}
 \end{subfigure}
\begin{subfigure}[t]{0.24\textwidth}
 \includegraphics[width=\textwidth]{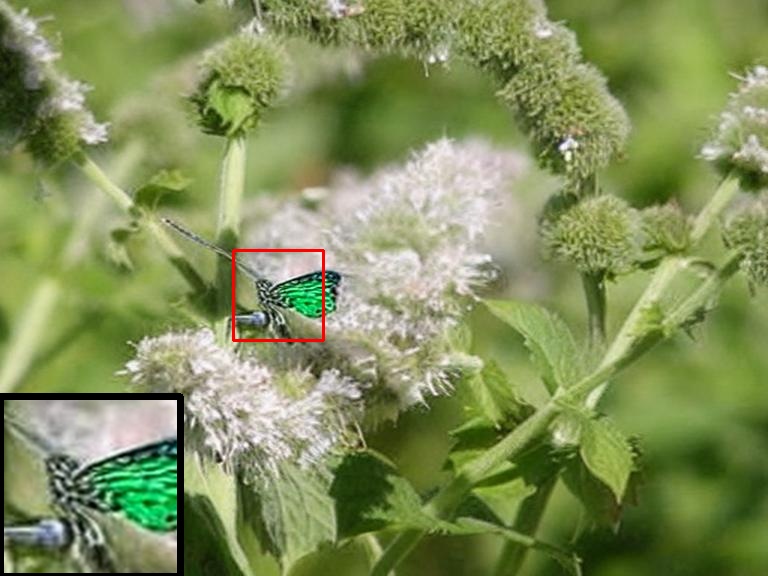}
 \end{subfigure}
\begin{subfigure}[t]{0.24\textwidth}
 \includegraphics[width=\textwidth]{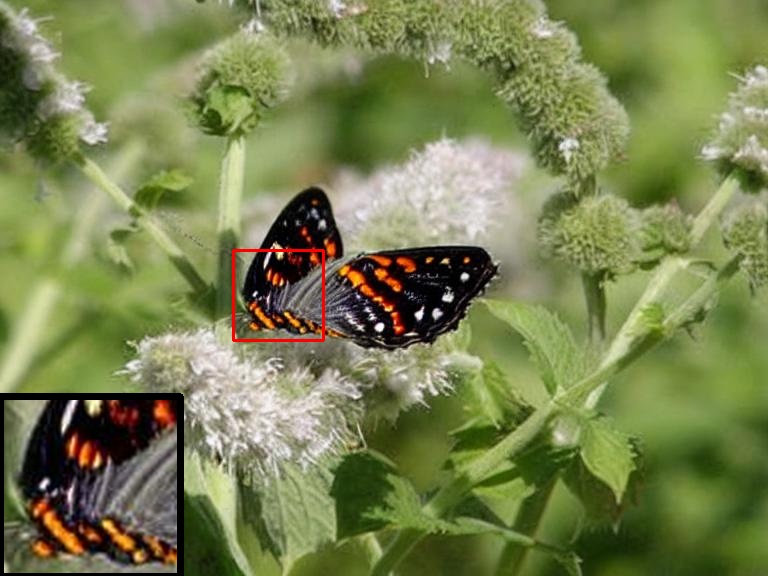}
 \end{subfigure}
\begin{subfigure}[t]{0.24\textwidth}
 \includegraphics[width=\textwidth]{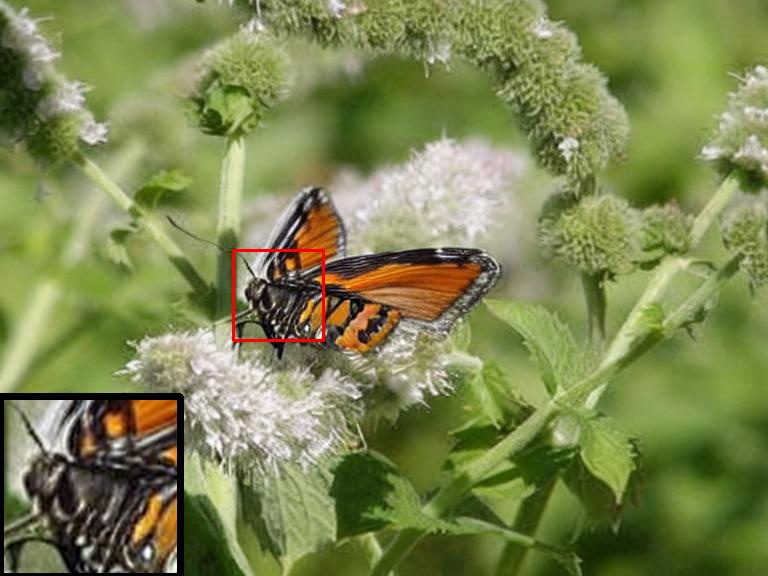}
 \end{subfigure}
\begin{subfigure}[t]{0.24\textwidth}
 \includegraphics[width=\textwidth]{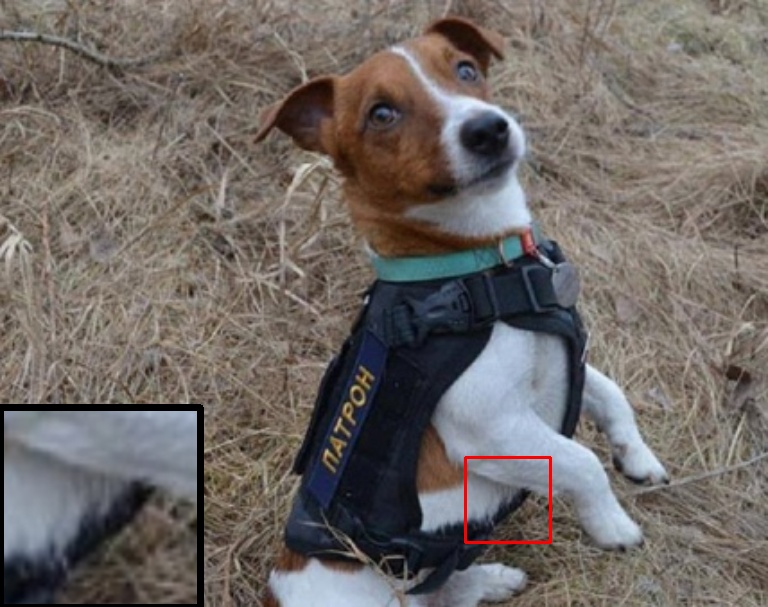}
 \caption{a) Original}
 \end{subfigure}
\begin{subfigure}[t]{0.24\textwidth}
 \includegraphics[width=\textwidth]{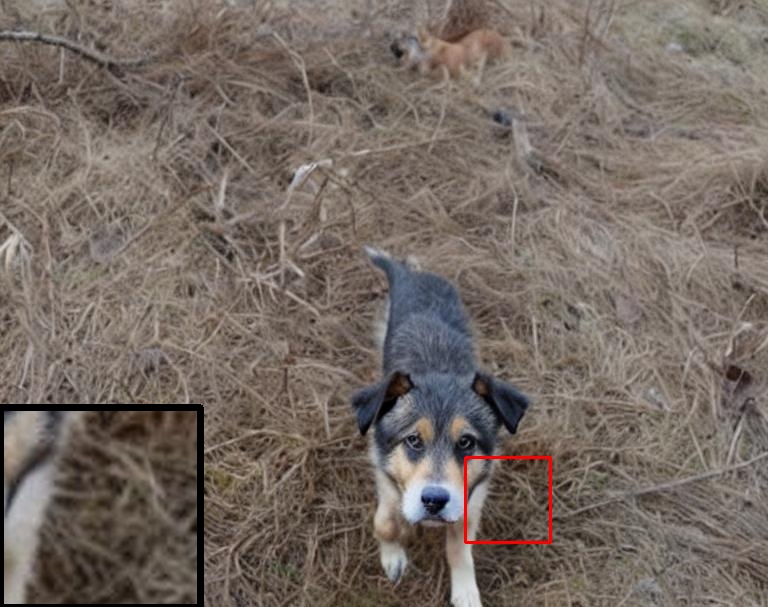}
 \caption{b) Inpainting}
 \end{subfigure}
\begin{subfigure}[t]{0.24\textwidth}
 \includegraphics[width=\textwidth]{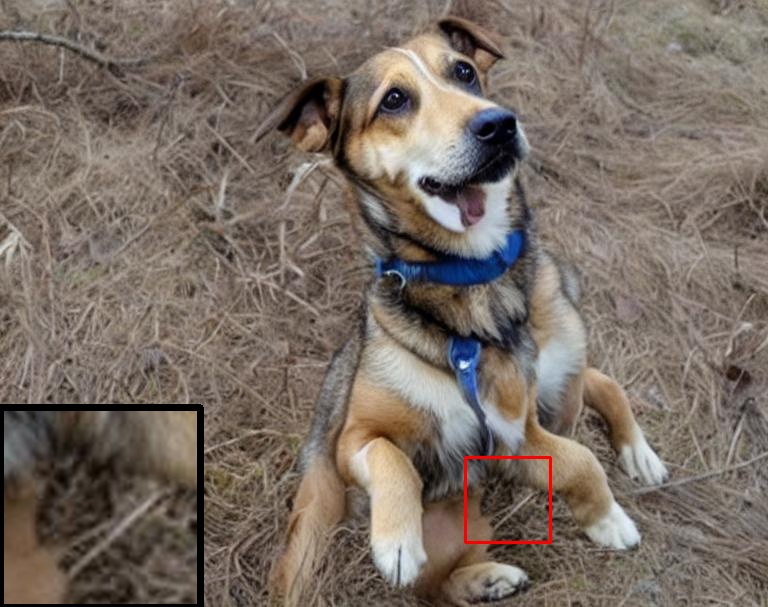}
 \caption{c) Depth Map}
 \end{subfigure}
\begin{subfigure}[t]{0.24\textwidth}
 \includegraphics[width=\textwidth]{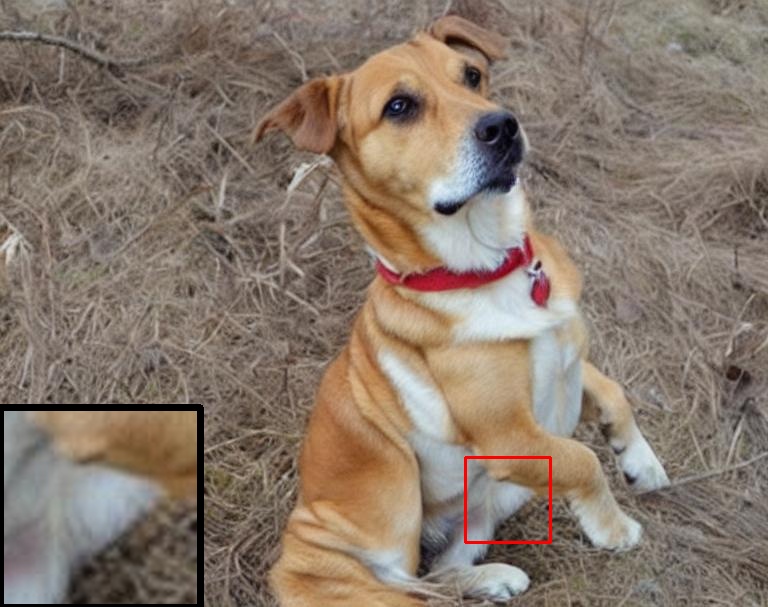}
 \caption{d) Depth + Edge}
 \end{subfigure}
\centering
\caption{\textbf{Inpainting Conditioning.} Inpainting without conditioning (b) does not preserve the structure of the object. Depth conditioning alone (c) fails to generate finer details and sharp edges. The full model (d) accurately preserves original annotations while producing high-quality samples.}
\label{fig:ablation}
\end{figure}

\subsection{Ablation Study}

\begin{table}[t]
\centering
\footnotesize
\begin{tabular}{lcccc}
\toprule
 \textbf{Component} & \textbf{$\boldsymbol{F_{max}}$}{$\uparrow$} & {\textbf{MAE}{$\downarrow$}} & $\boldsymbol{S_m}${$\uparrow$} & $\boldsymbol{F_{avg}}\uparrow$\\
\midrule
 \textit{full method} & \textbf{0.892} & \textbf{0.033} & \textbf{0.885} & \textbf{0.853}\\
 w/o instances (full image) & 0.889 & 0.035 & 0.881 & 0.848\\
 w/o edge and depth control & 0.886 & 0.036 & 0.881 & 0.846\\
 w/o edge control & 0.890 & 0.034 & 0.880 & 0.850\\
 w/o prompt engineering & 0.889 & 0.034 & 0.880 & 0.849 \\
 w/o mask refinement & 0.888 & 0.035 & 0.879 & 0.844\\
\bottomrule
\end{tabular}
\caption{ \textbf{Ablation study on DUTS.} Removing any component from the generation pipeline decreases the downstream task performance. The most important components are additional conditioning of the inpaiting model (ControlNet) and mask refinement.}
\label{t:ablation}
\end{table}
In this section, we verify the pipeline components' importance. We start with the full pipeline and analyze its performance on DUTS by subtracting individual components. We use TRACER \cite{tracer} with EfficientNet-0 \cite{effnet} for all experiments and report the results in \cref{t:ablation}.

Our method conditions the inpainting network on the predicted depth and edge map.
Removing it from the pipeline results in a performance drop across all metrics. 
We also train a model without the additional edge map input. This also results in reduced performance, indicating that both depth and edge map help preserve both high and low-level information. 
\Cref{fig:ablation} compares the original image to naive inpainting and control with depth, mask, and both. Additionally, we train a model on the version of the dataset with full image generated from only the depth and edges condition on the input. The main limitation of this method is still failing to generate scenes with a complex structure. 

\subsection{Prompt Engineering}\label{sec:exp:prompting}

\begin{figure}[t]
\captionsetup[subfigure]{font=tiny,labelformat=empty,labelfont={}}
\begin{subfigure}[t]{0.16\textwidth}
 \caption{cup}
 \includegraphics[width=\textwidth]{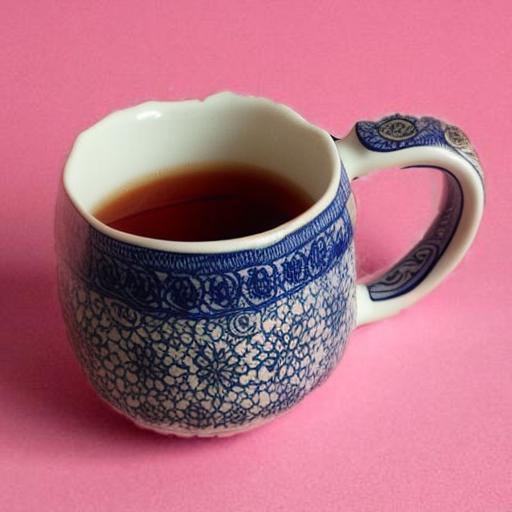}
\end{subfigure}
\begin{subfigure}[t]{0.16\textwidth}
 \caption{cup}
 \includegraphics[width=\textwidth]{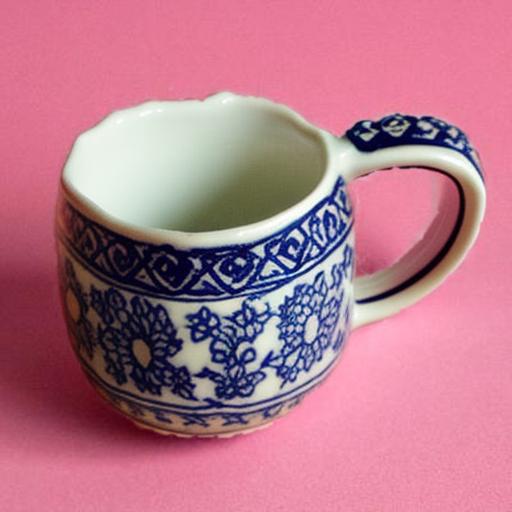}
\end{subfigure}
\begin{subfigure}[t]{0.16\textwidth}
 \caption{cup}
 \includegraphics[width=\textwidth]{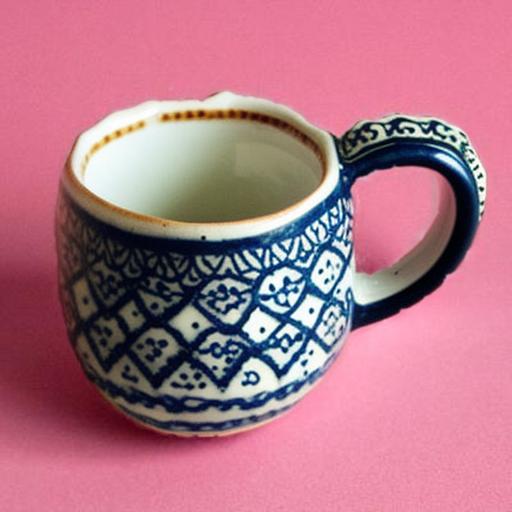}
\end{subfigure}
\begin{subfigure}[t]{0.16\textwidth}
 \caption{cup, blue, dark}
 \includegraphics[width=\textwidth]{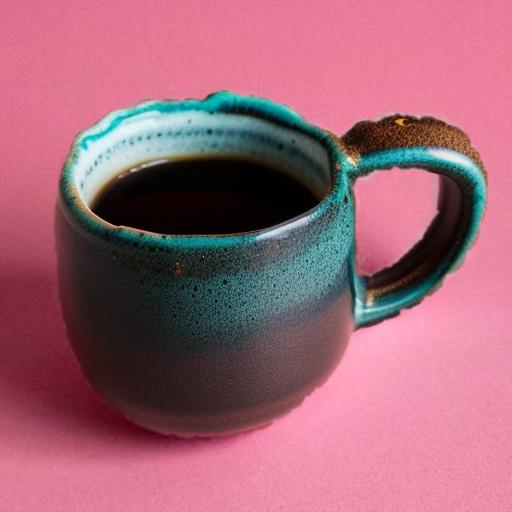}
\end{subfigure}
\begin{subfigure}[t]{0.16\textwidth}
 \caption{cup, yellow, sun}
 \includegraphics[width=\textwidth]{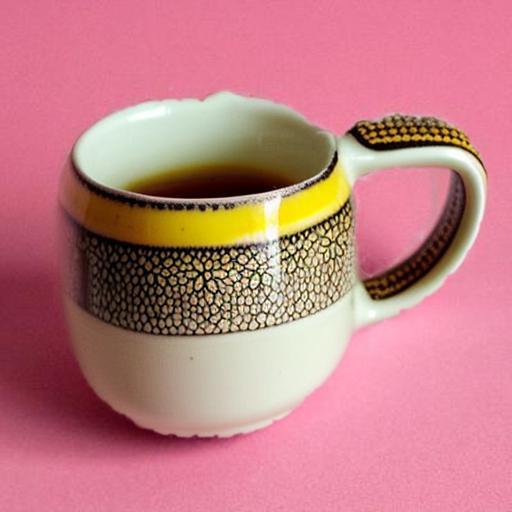}
\end{subfigure}
\begin{subfigure}[t]{0.16\textwidth}
 \caption{cup, light color}
 \includegraphics[width=\textwidth]{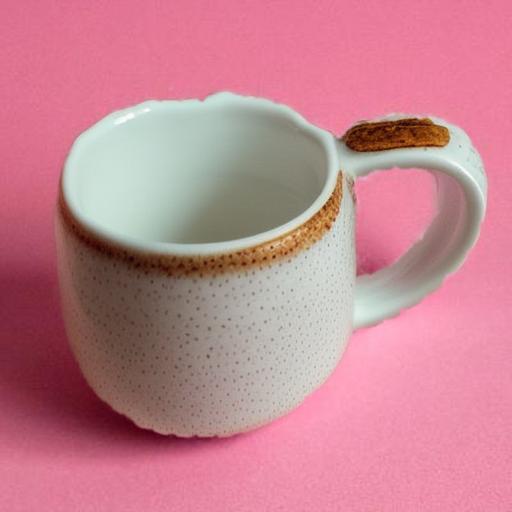}
\end{subfigure}
\centering
\caption{\textbf{Prompt Engineering.} Samples generated without (left) and with (right) color/lighting prompts. Extra prompts result in more diversity and variations.}
\label{fig:dataset_diversity}
\end{figure}

We introduce several improvements over simply using the class name as a prompt.
We verify this choice in \cref{t:ablation} where we show that the down-stream task performance decreases without ``prompt engineering''.
Further, we show visual examples in \cref{fig:dataset_diversity}.
The class label prompt limits the diversity of the generated cups, whereas including color and light information shows much more diverse results.
Finally, we observe that without including more detailed descriptions for people, the inpainter tends to sometimes fully remove (inpainting the background) complicated objects instead of redrawing them.

\textbf{Mask Refinement.} The pipeline performance is bounded by the performance of the underlying methods. In complex scenes, the error level of the depth or edge map model might cause the divergence between the generated image and the original annotations. The mask refinement module fixes disconnection, yet again improving the metrics and ensuring the stability of the generation pipeline.



\section{Conclusions}
\label{sec:conlusions}

In this paper, we introduce a method for object-level data augmentation. We combine a large pre-trained diffusion model with the low-level object representation to sample high-quality, diverse samples outside of the original dataset distribution. We demonstrate the method's efficiency as a data augmentation and anonymization technique. Additionally, we release the synthetic and anonymized versions of the standard detection benchmarks, COCO, VOC and DUTS.


\clearpage
\appendix

\section{Visual Examples} 
\begin{figure*}[htb]
\captionsetup[subfigure]{labelformat=empty}
\begin{subfigure}[t]{0.47\textwidth}
 \includegraphics[width=\textwidth]{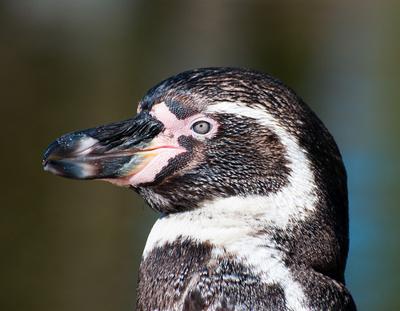}
 \end{subfigure}
\begin{subfigure}[t]{0.47\textwidth}
 \includegraphics[width=\textwidth]{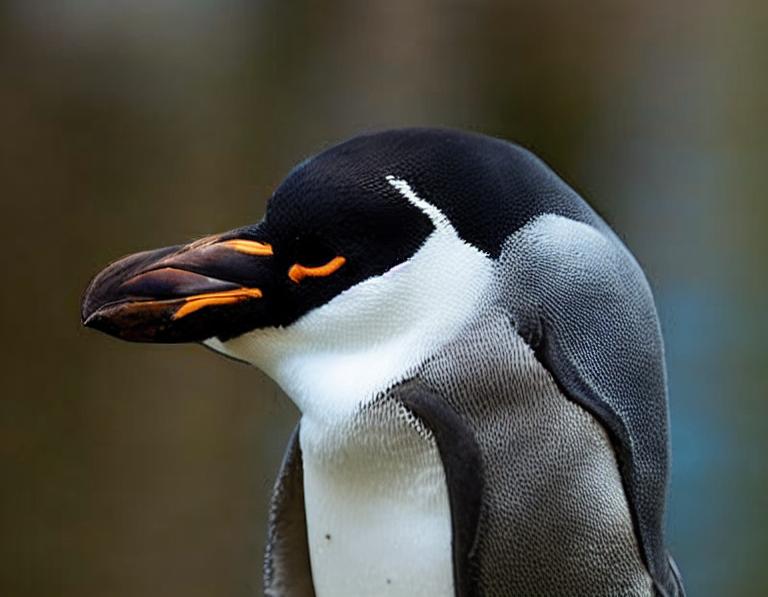}
 \end{subfigure}
\begin{subfigure}[t]{0.47\textwidth}
 \includegraphics[width=\textwidth]{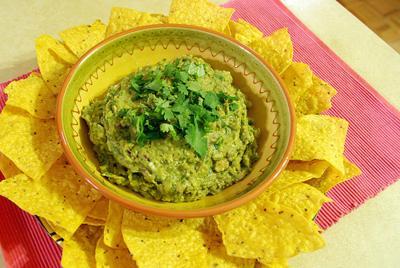}
 \end{subfigure}
\begin{subfigure}[t]{0.47\textwidth}
 \includegraphics[width=\textwidth]{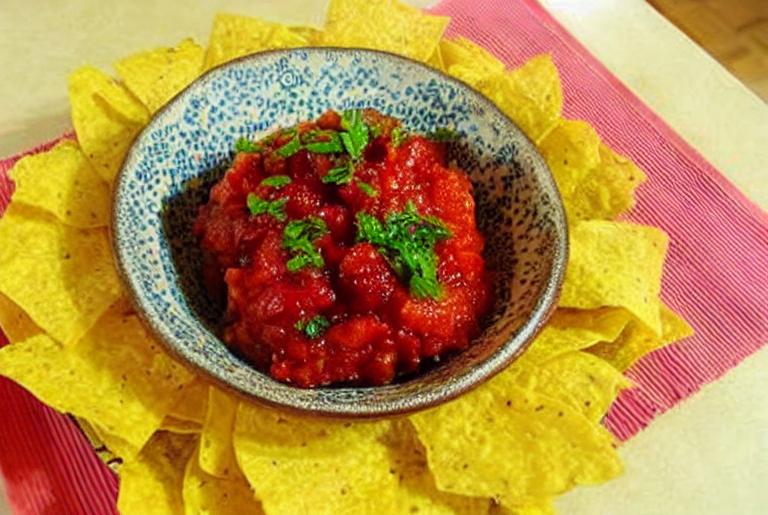}
 \end{subfigure}
\begin{subfigure}[t]{0.47\textwidth}
 \includegraphics[width=\textwidth]{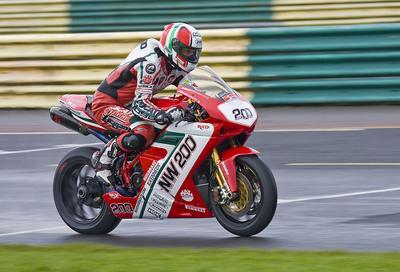}
 \end{subfigure}
\begin{subfigure}[t]{0.47\textwidth}
 \includegraphics[width=\textwidth]{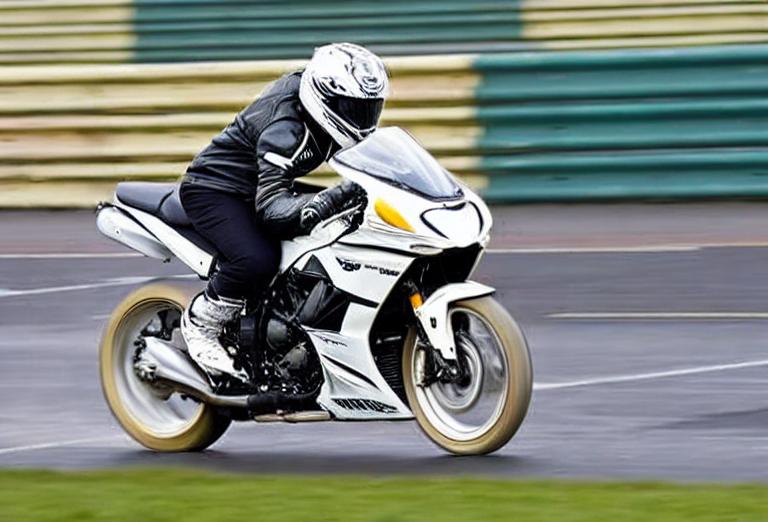}
 \end{subfigure}
\begin{subfigure}[t]{0.47\textwidth}
 \includegraphics[width=\textwidth]{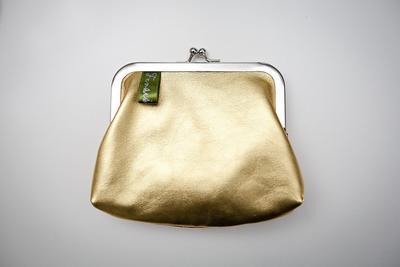}
 \caption{Original}
 \end{subfigure}
\begin{subfigure}[t]{0.47\textwidth}
 \includegraphics[width=\textwidth]{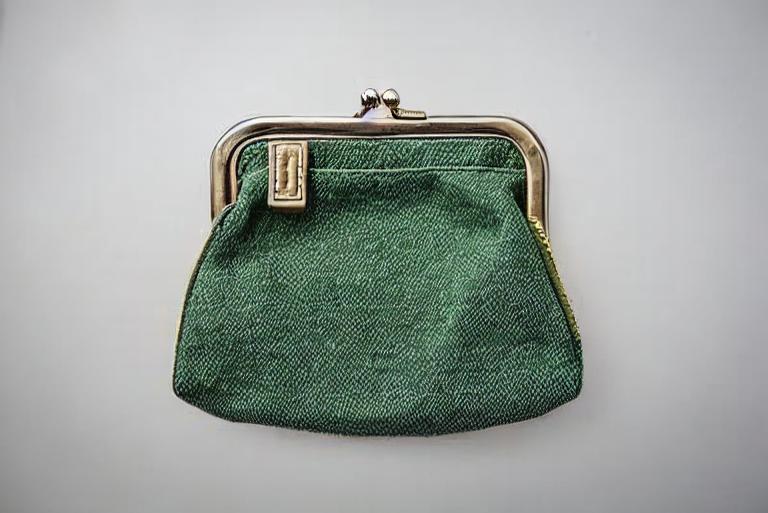}
 \caption{Augmented}
 \end{subfigure}
\centering
\caption{\textbf{Qualitative Examples.} Single Object Augmentation on DUTS.}
\label{fig:duts_examples}
\end{figure*}

\begin{figure*}[htb]
\captionsetup[subfigure]{labelformat=empty}
\begin{subfigure}[t]{0.47\textwidth}
 \includegraphics[width=\textwidth]{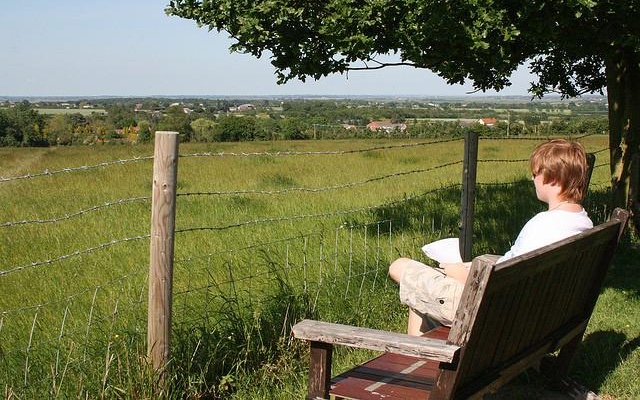}
 \end{subfigure}
\begin{subfigure}[t]{0.47\textwidth}
 \includegraphics[width=\textwidth]{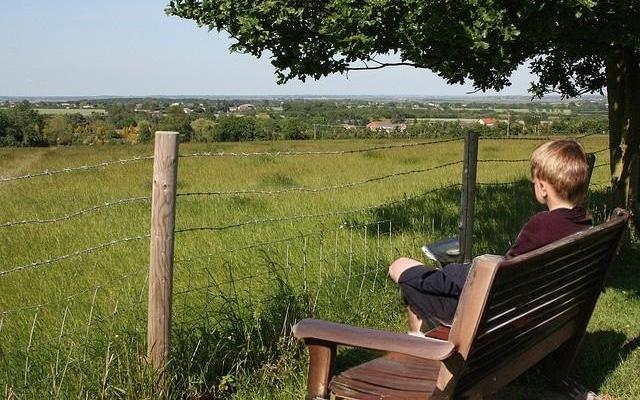}
 \end{subfigure}
\begin{subfigure}[t]{0.47\textwidth}
 \includegraphics[width=\textwidth]{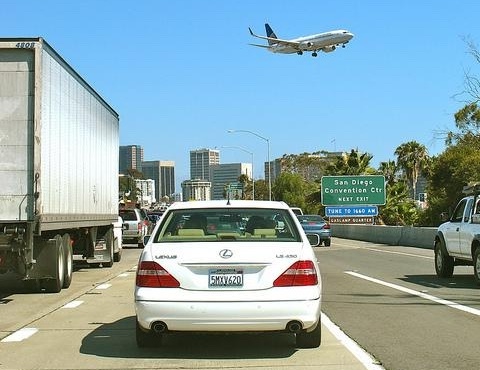}
 \end{subfigure}
\begin{subfigure}[t]{0.47\textwidth}
 \includegraphics[width=\textwidth]{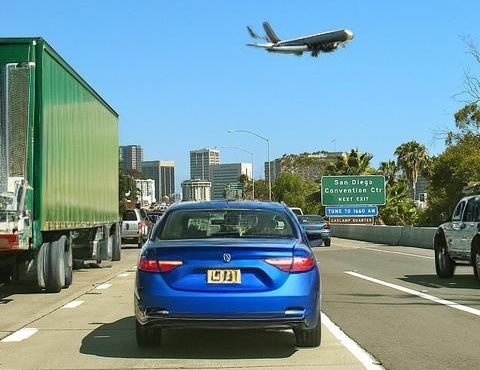}
 \end{subfigure}
\begin{subfigure}[t]{0.47\textwidth}
 \includegraphics[width=\textwidth]{images/supplementary/train_56860_0.jpg}
 \end{subfigure}
\begin{subfigure}[t]{0.47\textwidth}
 \includegraphics[width=\textwidth]{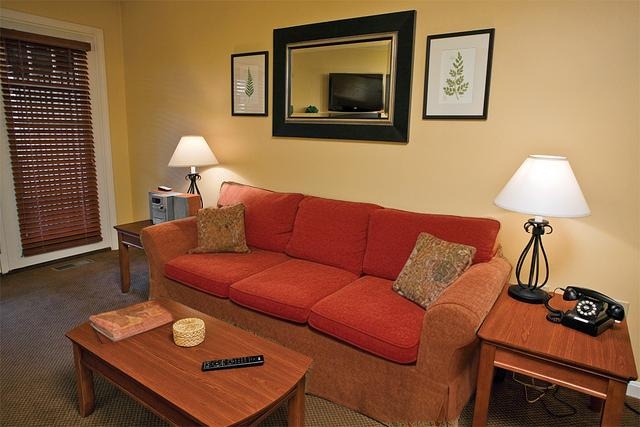}
 \end{subfigure}
\begin{subfigure}[t]{0.47\textwidth}
 \includegraphics[width=\textwidth]{images/supplementary/train_55433_0.jpg}
 \caption{Original}
 \end{subfigure}
\begin{subfigure}[t]{0.47\textwidth}
 \includegraphics[width=\textwidth]{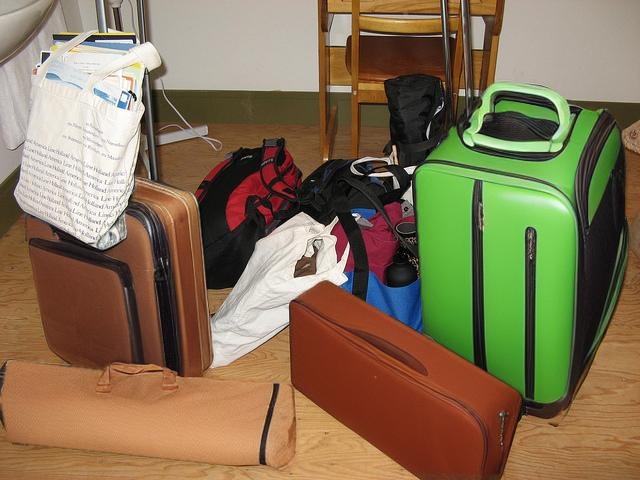}
 \caption{Augmented}
 \end{subfigure}
\centering
\caption{\textbf{Complex Scenes.} Mixing real and generated objects.}
\label{fig:coco_examples}
\end{figure*}

\begin{figure*}[htb]
\captionsetup[subfigure]{labelformat=empty}
\begin{subfigure}[t]{0.24\textwidth}
 \includegraphics[width=\textwidth]{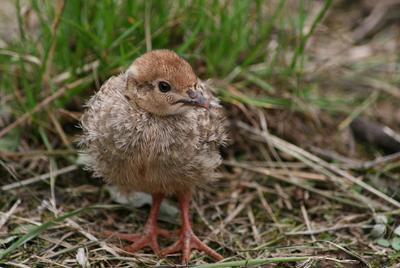}
 \end{subfigure}
\begin{subfigure}[t]{0.24\textwidth}
 \includegraphics[width=\textwidth]{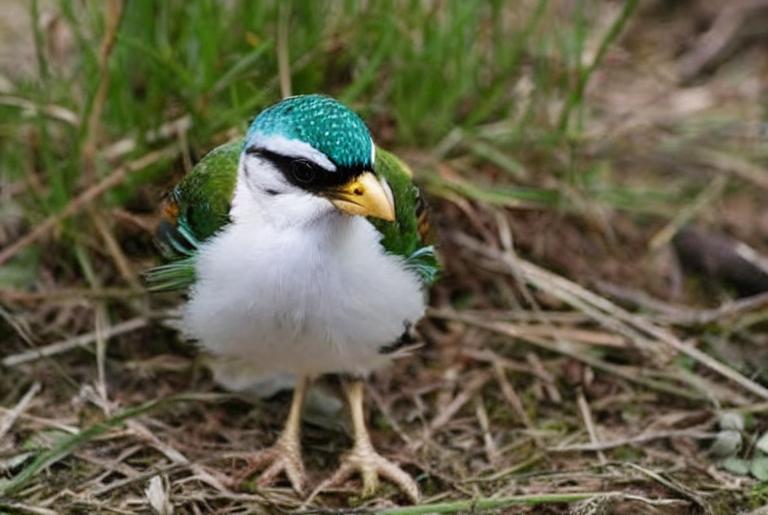}
 \end{subfigure}
\begin{subfigure}[t]{0.24\textwidth}
 \includegraphics[width=\textwidth]{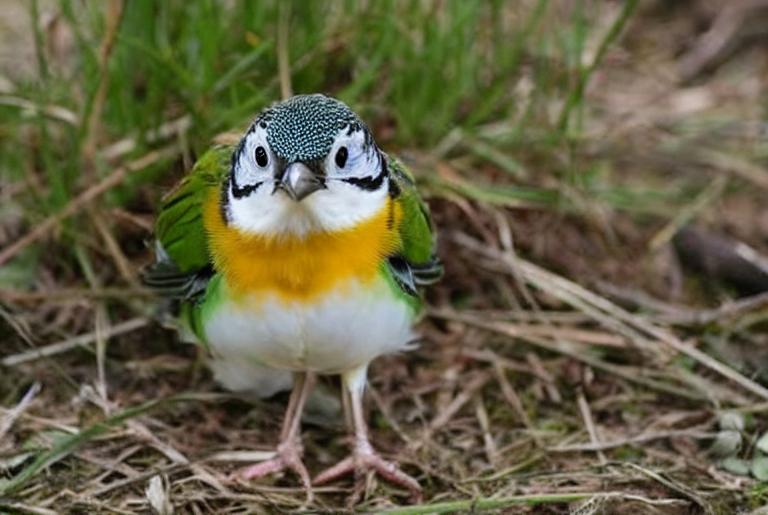}
 \end{subfigure}
\begin{subfigure}[t]{0.24\textwidth}
 \includegraphics[width=\textwidth]{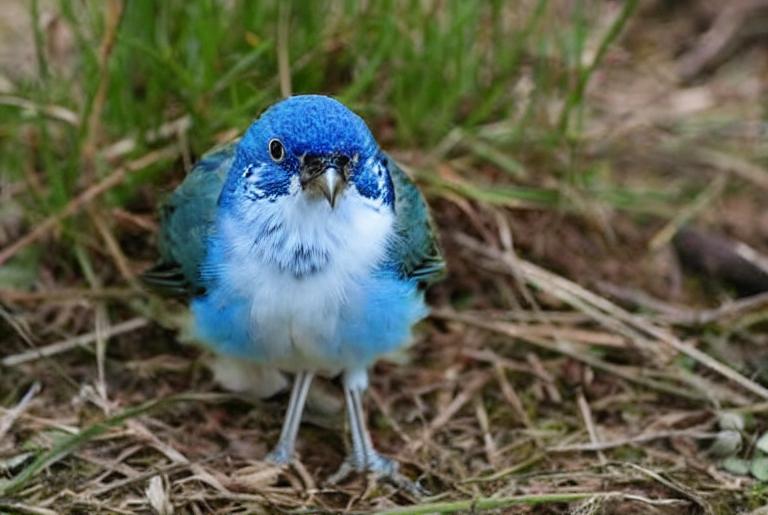}
 \end{subfigure}
\begin{subfigure}[t]{0.24\textwidth}
 \includegraphics[width=\textwidth]{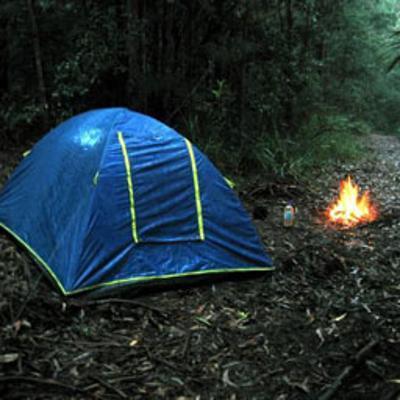}
 \end{subfigure}
\begin{subfigure}[t]{0.24\textwidth}
 \includegraphics[width=\textwidth]{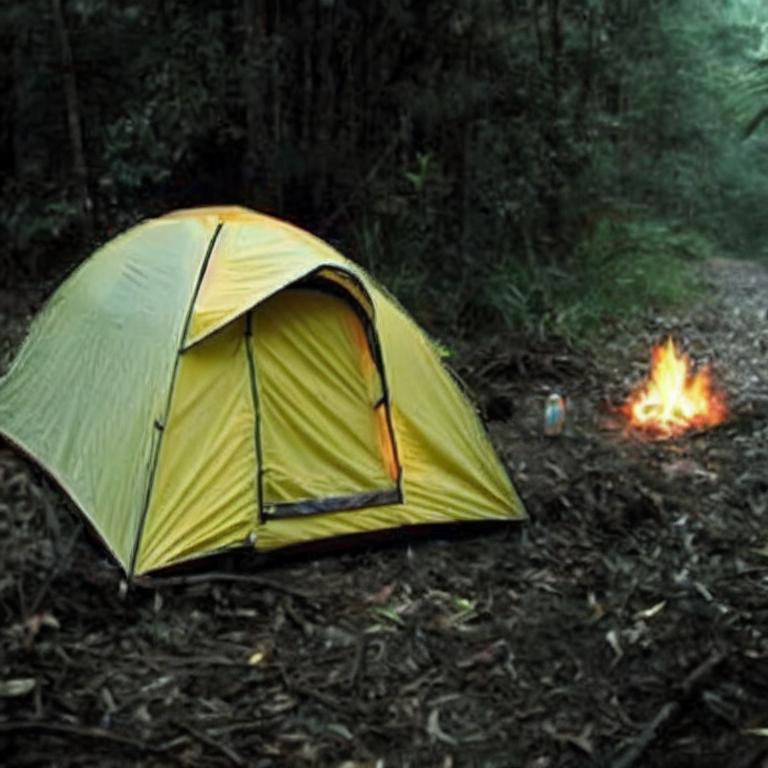}
 \end{subfigure}
\begin{subfigure}[t]{0.24\textwidth}
 \includegraphics[width=\textwidth]{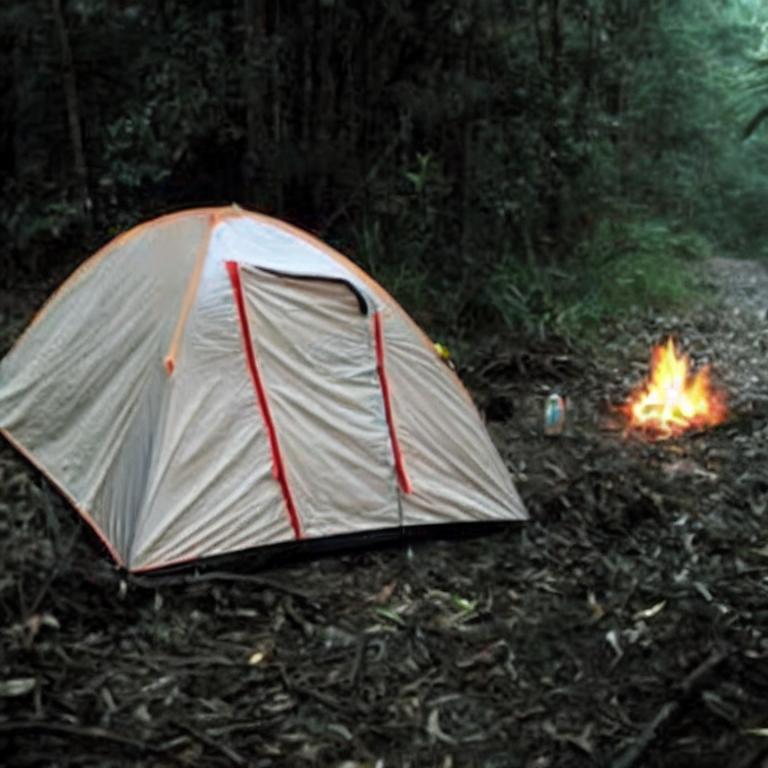}
 \end{subfigure}
\begin{subfigure}[t]{0.24\textwidth}
 \includegraphics[width=\textwidth]{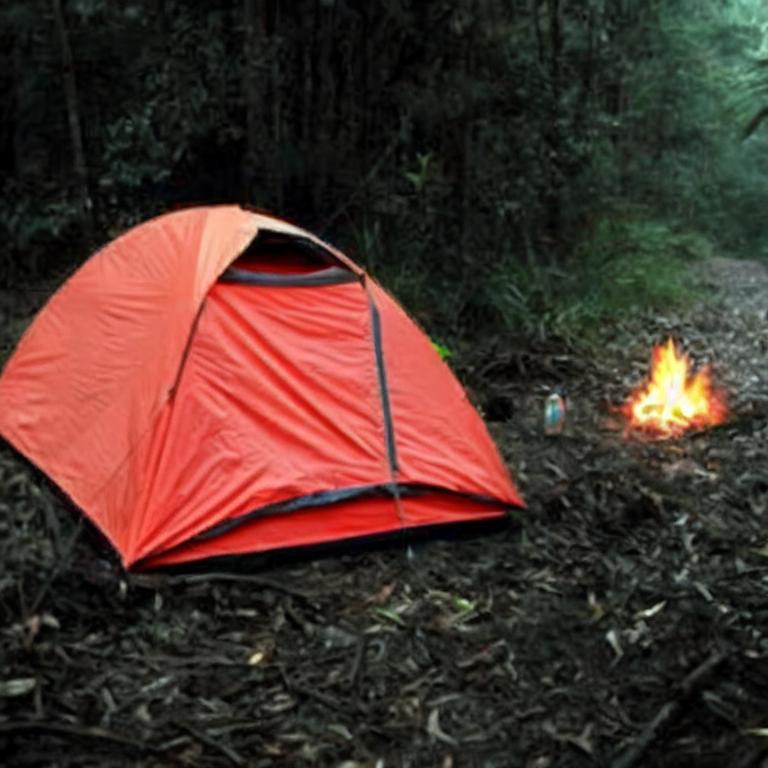}
 \end{subfigure}
 \begin{subfigure}[t]{0.24\textwidth}
 \includegraphics[width=\textwidth]{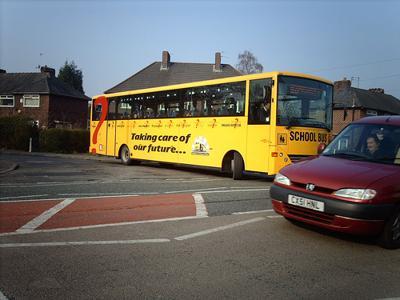}
 \end{subfigure}
\begin{subfigure}[t]{0.24\textwidth}
 \includegraphics[width=\textwidth]{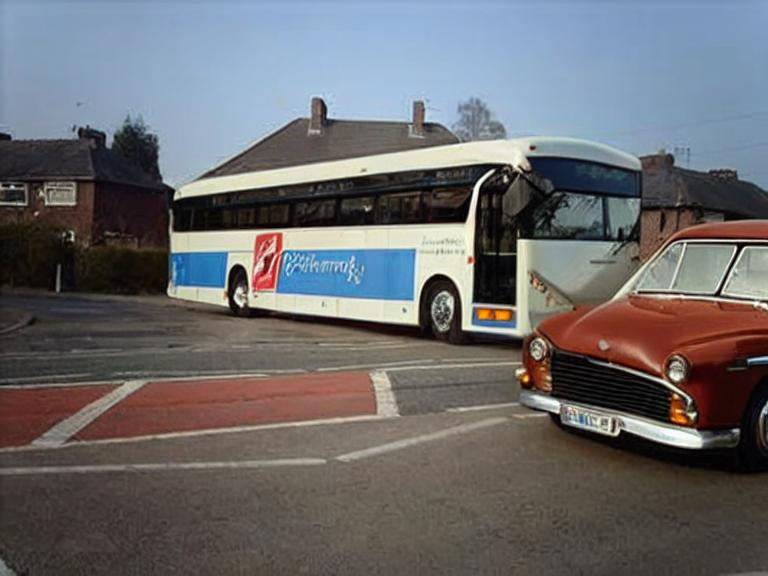}
 \end{subfigure}
\begin{subfigure}[t]{0.24\textwidth}
 \includegraphics[width=\textwidth]{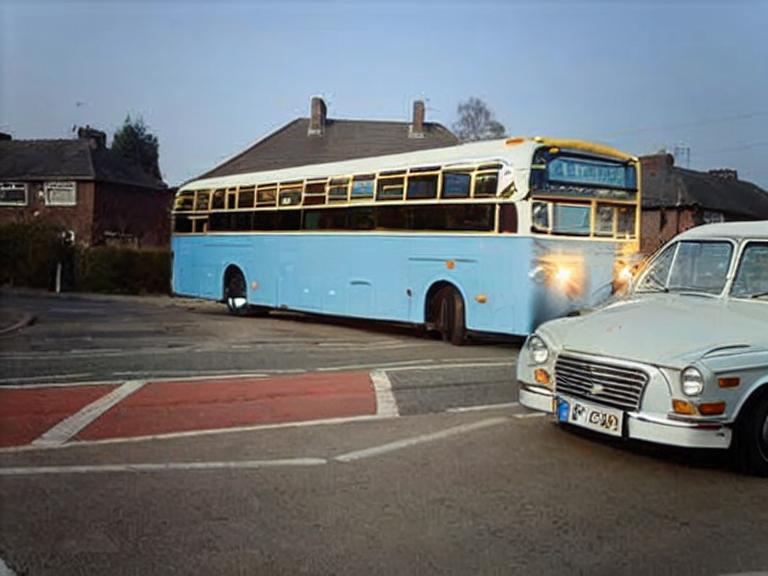}
 \end{subfigure}
\begin{subfigure}[t]{0.24\textwidth}
 \includegraphics[width=\textwidth]{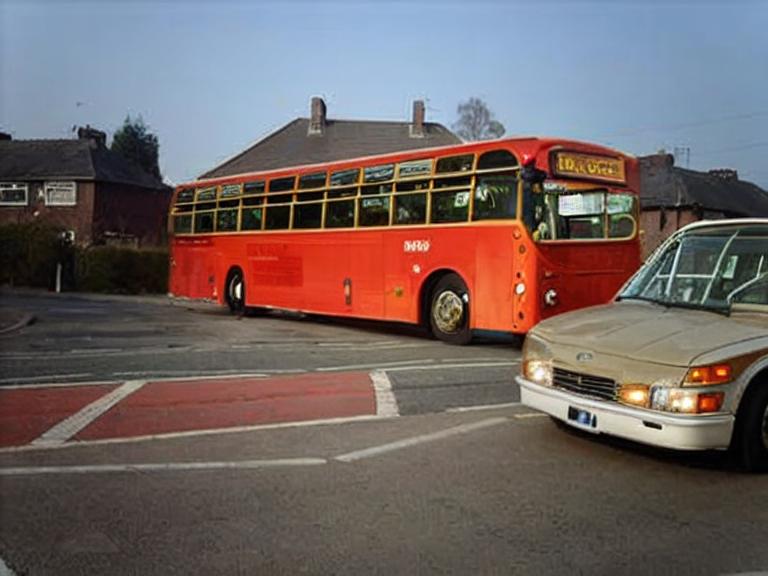}
 \end{subfigure}
 \begin{subfigure}[t]{0.24\textwidth}
 \includegraphics[width=\textwidth]{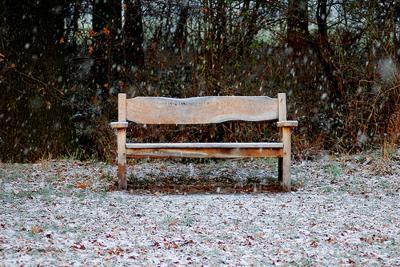}
 \end{subfigure}
\begin{subfigure}[t]{0.24\textwidth}
 \includegraphics[width=\textwidth]{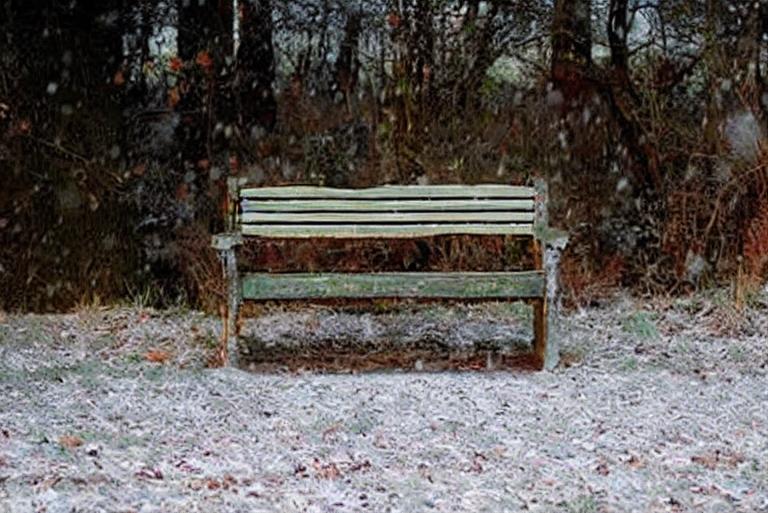}
 \end{subfigure}
\begin{subfigure}[t]{0.24\textwidth}
 \includegraphics[width=\textwidth]{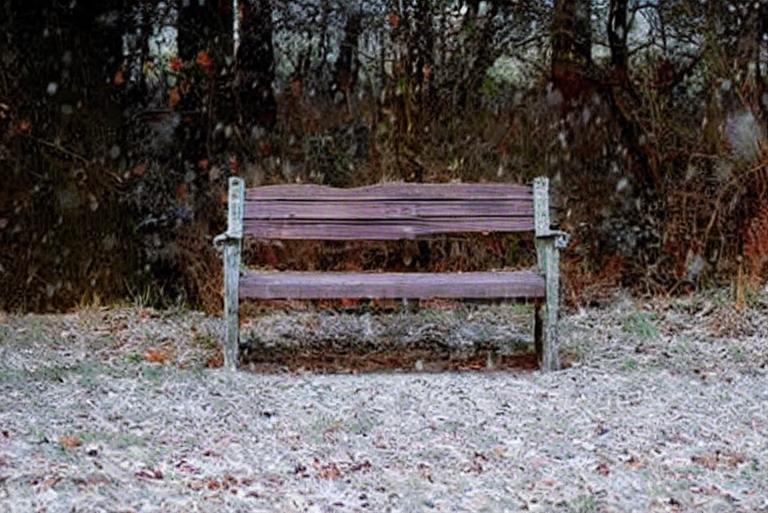}
 \end{subfigure}
\begin{subfigure}[t]{0.24\textwidth}
 \includegraphics[width=\textwidth]{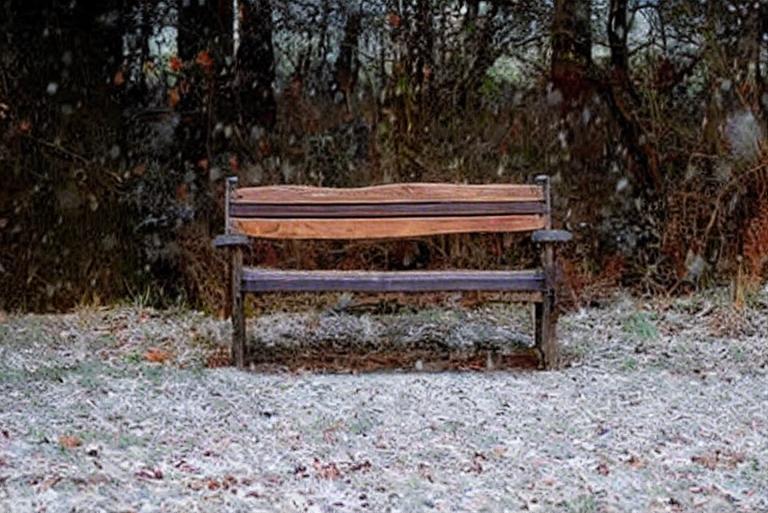}
 \end{subfigure}
 \begin{subfigure}[t]{0.24\textwidth}
 \includegraphics[width=\textwidth]{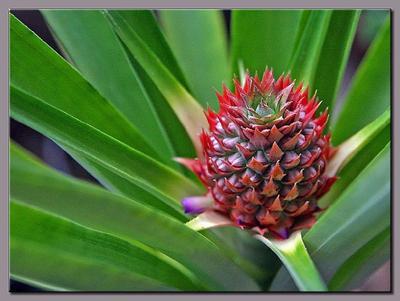}
 \end{subfigure}
\begin{subfigure}[t]{0.24\textwidth}
 \includegraphics[width=\textwidth]{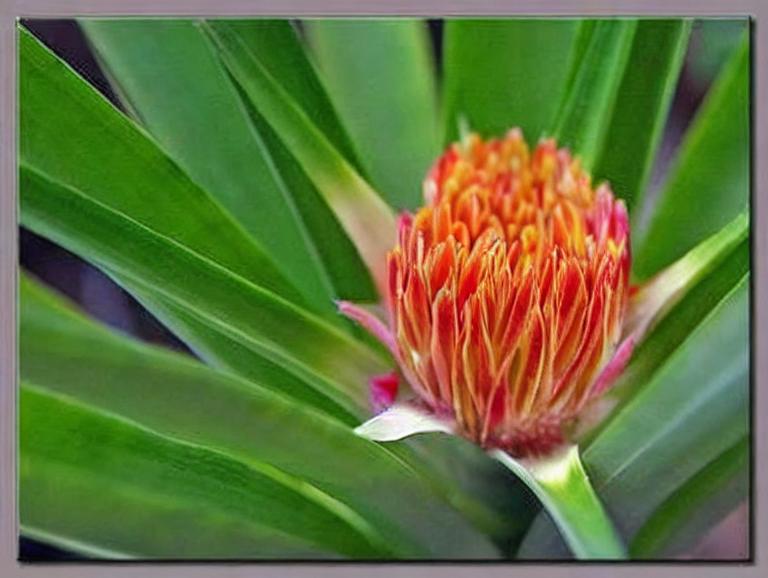}
 \end{subfigure}
\begin{subfigure}[t]{0.24\textwidth}
 \includegraphics[width=\textwidth]{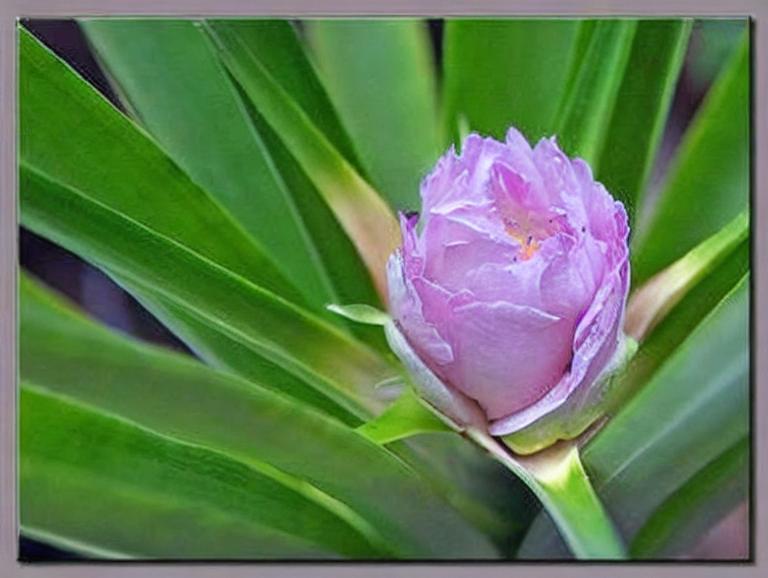}
 \end{subfigure}
\begin{subfigure}[t]{0.24\textwidth}
 \includegraphics[width=\textwidth]{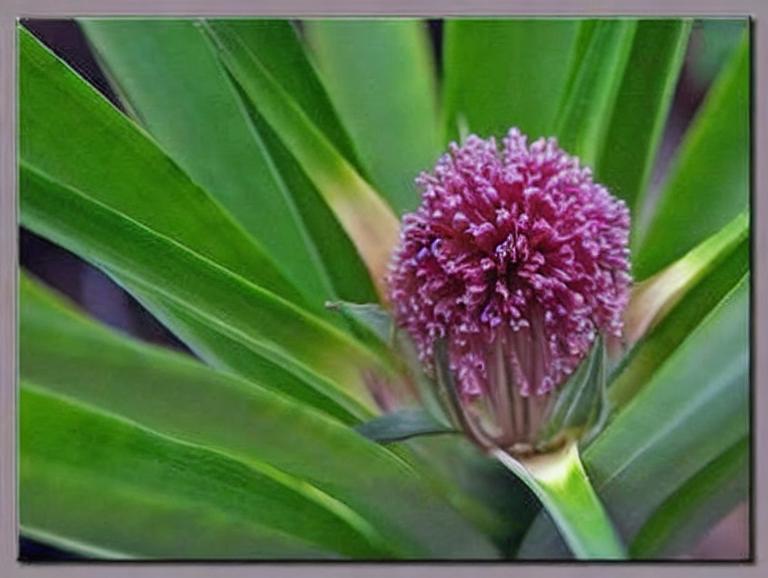}
 \end{subfigure}
  \begin{subfigure}[t]{0.24\textwidth}
 \includegraphics[width=\textwidth]{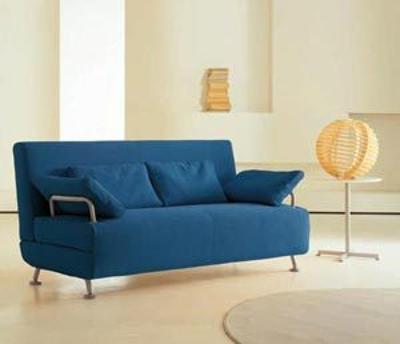}
 \end{subfigure}
\begin{subfigure}[t]{0.24\textwidth}
 \includegraphics[width=\textwidth]{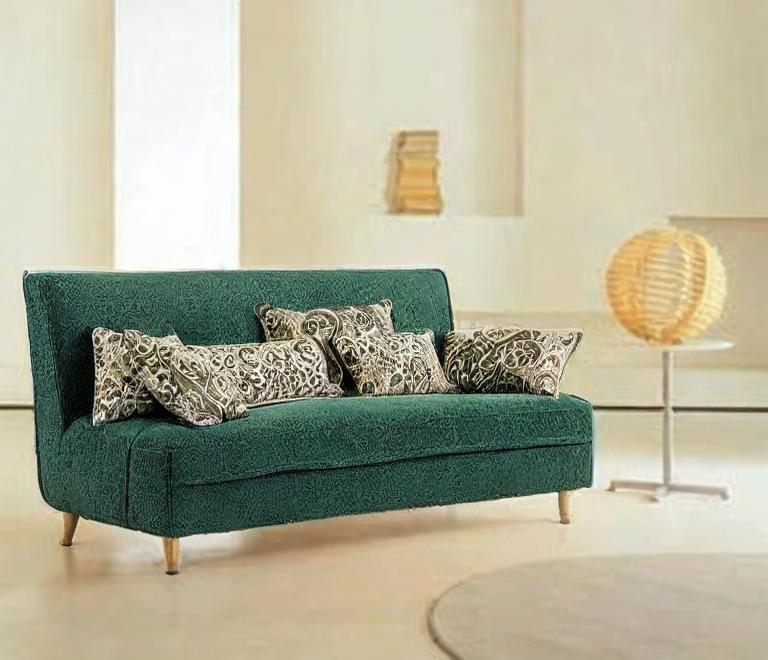}
 \end{subfigure}
\begin{subfigure}[t]{0.24\textwidth}
 \includegraphics[width=\textwidth]{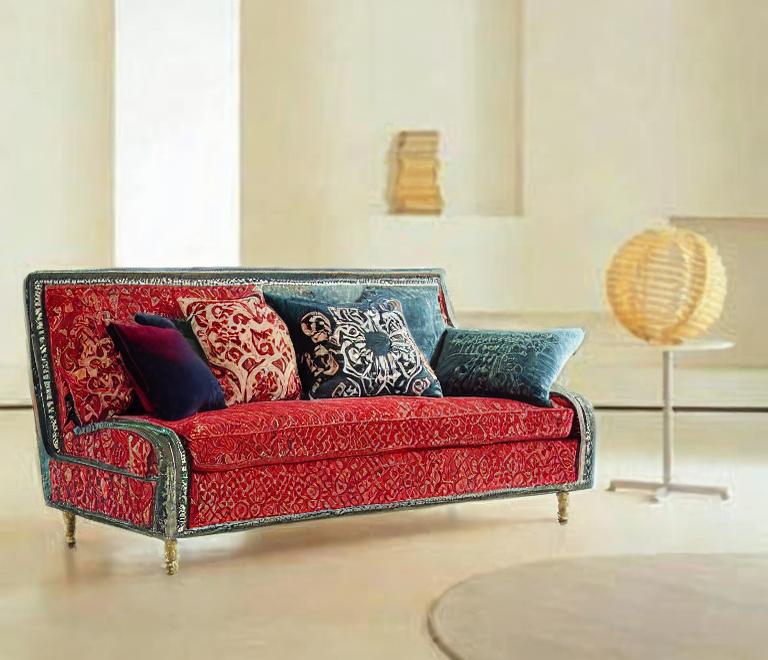}
 \end{subfigure}
\begin{subfigure}[t]{0.24\textwidth}
 \includegraphics[width=\textwidth]{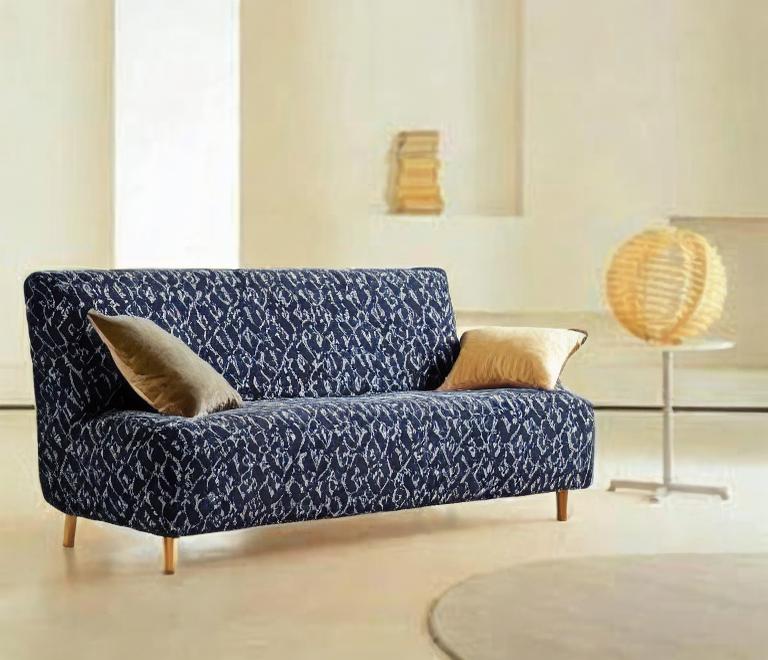}
 \end{subfigure}
\centering
\caption{\textbf{Diversity.} Showing three generated examples from the original (left) on DUTS.}
\label{fig:diversity_examples}
\end{figure*}

We include more visual examples of the method performance for scenes with one main object (\cf \cref{fig:duts_examples}) and complex scenes combining real and generated objects (\cf \cref{fig:coco_examples}). 
Additionally, we demonstrate visually diverse samples on different object categories (\cf \cref{fig:diversity_examples}).

\section{Data Anonymization} 
To measure the efficiency of the method to anonymize sensitive data we check how often a replaced face can be matched with its original appearance using a face identification model. 
To this end, we use ArcFace \cite{arcface} on the pairs of original and generated images from the COCO dataset, validating whether people can be re-identified after applying our method. 
As the model does not work on children, manually filter them out from the dataset,
From 64115 images with people and 262465 faces in them, only 373 (0.14\%) we re-indentified. 
Inspecting these cases (\cref{fig:reid_examples}), we find that they are almost exclusively false positives of the ReID model.

\section{Visual Quality} 
We measure the impact of our method's components on the visual quality of the generated images in \cref{t:ablation_fid}.
We calculate FID \cite{fid} and Inception Score \cite{inception_score} for every setup to measure both the diversity and realism of the generated samples.
We find that constraining the model with edges and depth map has a small impact on the FID score, likely
because more constraints inherently limit the diversity of the generated images. 
\begin{table}[t!]
\footnotesize
\setlength{\tabcolsep}{1.5pt}
\centering
\begin{tabular}{lccc}
\toprule
 \textbf{Component} & \textbf{FID}{$\downarrow$} & {\textbf{Inception Score}{$\uparrow$}} & \textbf{$\boldsymbol{F_{max}}$}{$\uparrow$}\\
\midrule
 \textit{full method} & 1.10 & \textbf{50.86} & \textbf{0.892}\\
 w/o edge and depth control & \textbf{0.92} & 38.00 & 0.886\\
 w/o edge control  & 1.41 & 37.35 & 0.890\\
 w/o prompt engineering & 1.76 & 38.31 & 0.889\\
\bottomrule

\end{tabular}
\caption{ \textbf{Visual Quality.} Without edge- and depth control the method generates images of high quality. They do not, however, match the original annotations. Removing edge map guidance and prompt engineering significantly drops both quality and diversity.}
\label{t:ablation_fid}
\end{table}
\begin{table*}[h!]
\footnotesize
\centering
\resizebox{\linewidth}{!}{\begin{tabular}{lc|ccc|ccc}
\toprule
& & \multicolumn{3}{c}{all (excl.~people)} & \multicolumn{3}{c}{all (excl.~cars)}\\
Model & Dataset & AP & AP$_{50}$ & AP$_{75}$ & AP & AP$_{50}$ & AP$_{75}$\\
\midrule
Deformable-DETR & Real & \textbf{39.3} & \textbf{60.1} & \textbf{41.5} & \textbf{38.5} & \textbf{59.9} & 40.4 \\
Deformable-DETR & Anonymized People & 38.7 & 59.1 & 40.9 & - & - & - \\
Deformable-DETR & Anonymized Cars & - & - & - & 38.2 & 59.5 & \textbf{40.6} \\
\bottomrule
\end{tabular}}
\caption{ \textbf{Data Anonymization.} Fully replacing a category with synthetic examples has almost no impact on the performance of other classes.}
\label{t:anonym}
\end{table*}

\section{Limitations and Future Work}

\begin{figure}[htb]
\captionsetup[subfigure]{labelformat=empty}
\begin{subfigure}[t]{0.32\linewidth}
\includegraphics[width=\textwidth]{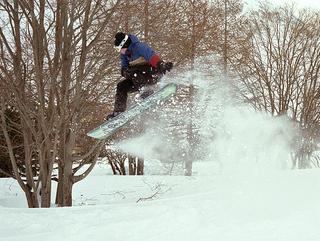}
\caption{Original}
\end{subfigure}
\begin{subfigure}[t]{0.32\linewidth}
\includegraphics[width=\textwidth]{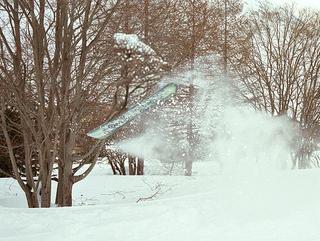}
\caption{``person''}
\end{subfigure}
\begin{subfigure}[t]{0.32\linewidth}
\includegraphics[width=\textwidth]{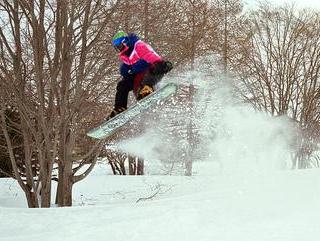}
\caption{``person snowboarding''}
\end{subfigure}
\centering
\caption{\textbf{Person Prompts.} Providing a more detailed text description for people improves the stability of the method. }
\label{fig:model_fail}
\end{figure}

\begin{figure}[htb]
 \captionsetup[subfigure]{labelformat=empty}
 \begin{subfigure}[t]{0.32\textwidth}
 \includegraphics[width=\textwidth]{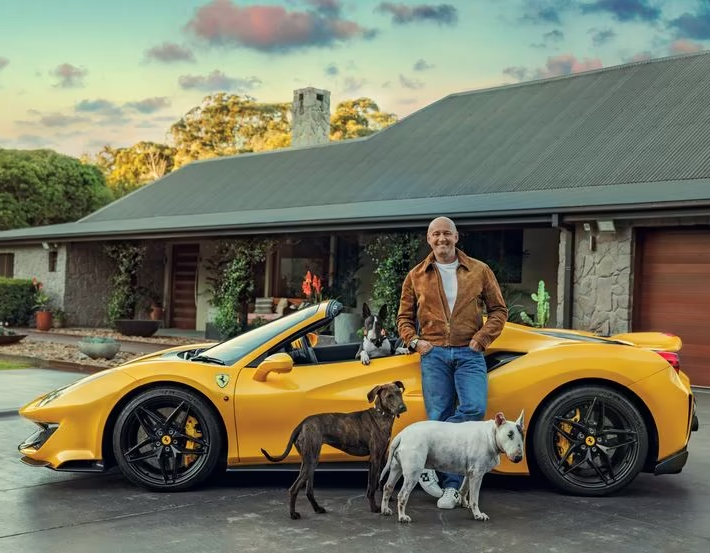}
 \caption{Original}
 \end{subfigure}
\begin{subfigure}[t]{0.32\textwidth}
 \includegraphics[width=\textwidth]{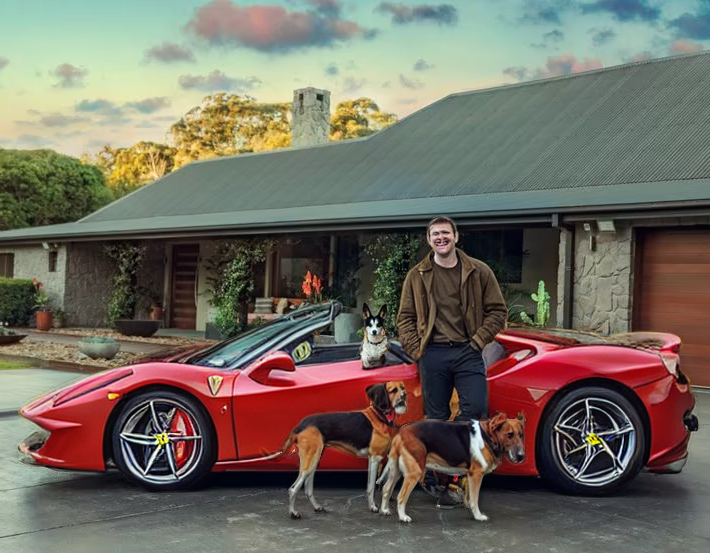}
 \caption{Stable Diffusion 1.5}
 \end{subfigure}
\begin{subfigure}[t]{0.32\textwidth}
 \includegraphics[width=\textwidth]{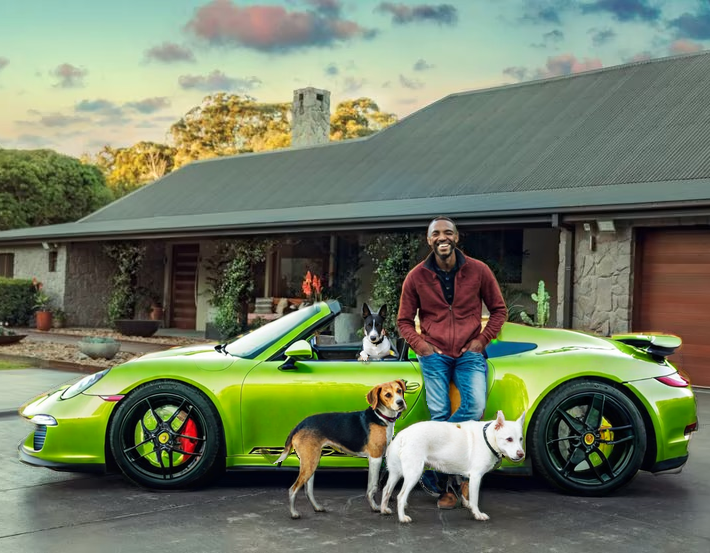}
 \caption{Stable Diffusion XL}
 \end{subfigure}
\centering
\caption{\textbf{Stronger Generative Model - Better Performance.} Recent advancements in image generation models open new possibilities to utilize synthetic data for vision tasks. In our pipeline, more advanced diffusion models generate more realistic and detailed objects in higher resolution.}
\label{fig:sdxl}
\end{figure}

\begin{figure*}[htb]
\captionsetup[subfigure]{labelformat=empty}
\begin{subfigure}[t]{0.48\textwidth}
 \includegraphics[width=\textwidth]{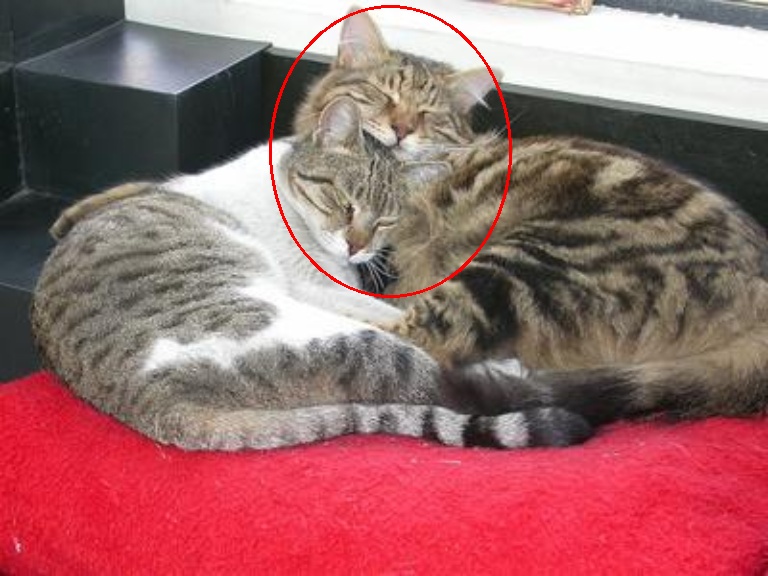}
 \end{subfigure}
\begin{subfigure}[t]{0.48\textwidth}
 \includegraphics[width=\textwidth]{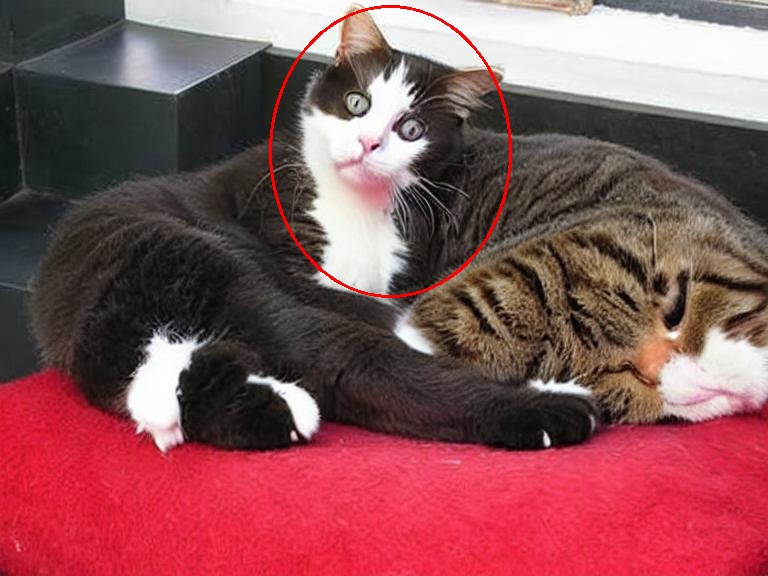}
 \end{subfigure}
 \begin{subfigure}[t]{0.48\textwidth}
 \includegraphics[width=\textwidth]{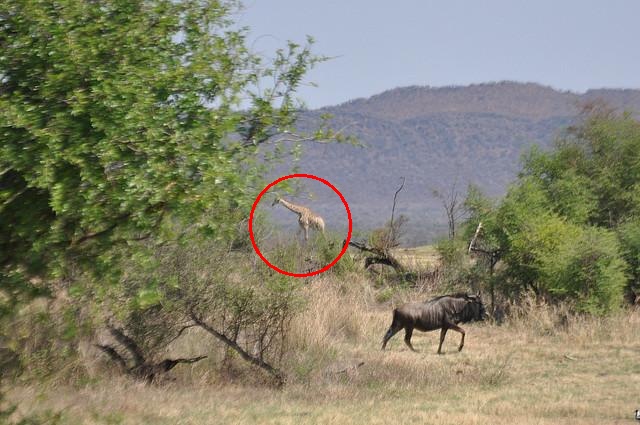}
 \end{subfigure}
\begin{subfigure}[t]{0.48\textwidth}
 \includegraphics[width=\textwidth]{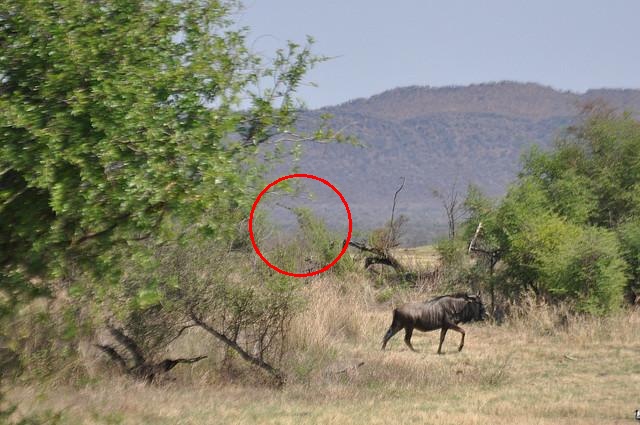}
 \end{subfigure}
 \begin{subfigure}[t]{0.48\textwidth}
 \includegraphics[width=\textwidth]{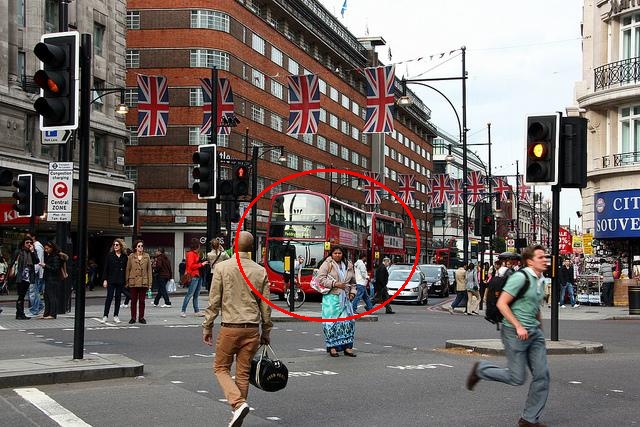}
 \end{subfigure}
\begin{subfigure}[t]{0.48\textwidth}
 \includegraphics[width=\textwidth]{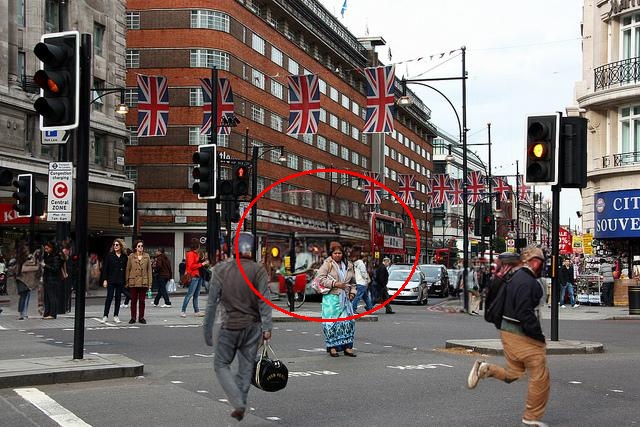}
 \end{subfigure}
\begin{subfigure}[t]{0.48\textwidth}
 \includegraphics[width=\textwidth]{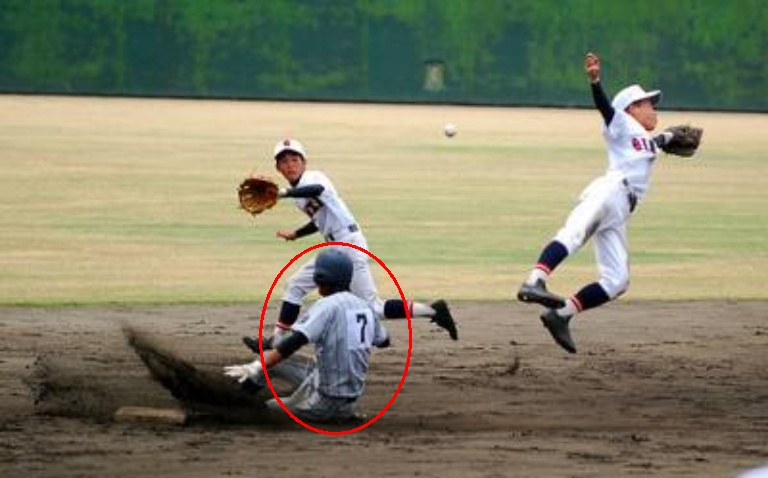}
 \caption{Original}
 \end{subfigure}
\begin{subfigure}[t]{0.48\textwidth}
 \includegraphics[width=\textwidth]{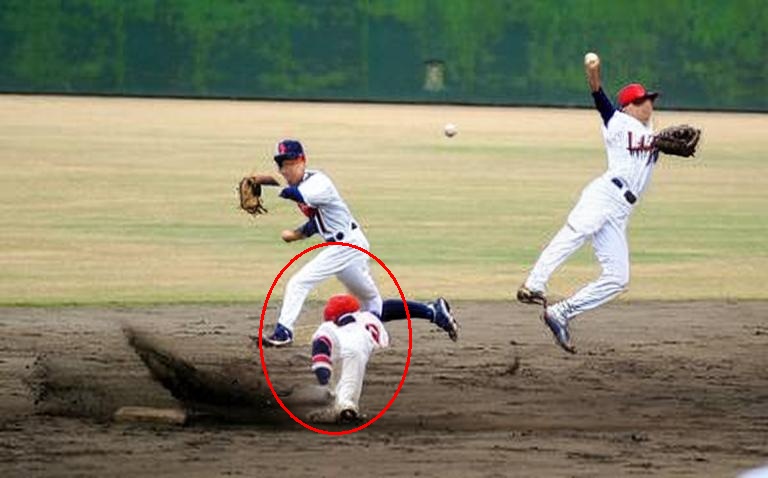}
 \caption{Generated}
 \end{subfigure}
\centering
\caption{\textbf{Failure Cases.} Examples of removed objects and lower-quality generations. Red ellipses mark objects that were not repainted correctly.}
\label{fig:fail_cases}
\end{figure*}

The method's performance is bounded by the performance of the main components: the inpainting and the control net model.
For small and occluded objects, the inpainting model sometimes completely removes the object and inpaints with the background (\cf \cref{fig:fail_cases}). \Cref{fig:model_fail} shows the impact of prompt engineering for the person category on the inpainting.
Recent advancements in image generation \cite{sdxl} could help mitigate this as well as further close the gap with the real-world data distribution. The pipeline is invariant to the choice of generative model, so employing more advanced generative models has the potential to further improve performance (\cf \cref{fig:sdxl}).
Further, the method is limited to the domain of the large text to mage generative model (real-world scenes) and thus can not be applied to specific domains such as satellite or medical images.

\begin{table*}[h!]
\footnotesize
\centering
\setlength{\tabcolsep}{0.5pt}
\resizebox{\linewidth}{!}{\begin{tabular}{lc|cccc|cccc|cccc|cccc|cccc}
\toprule
& & \multicolumn{4}{c}{ECCSD} & \multicolumn{4}{c}{DUTS-TE} & \multicolumn{4}{c}{DUT-OMRON} & \multicolumn{4}{c}{HKU-IS} & \multicolumn{4}{c}{HRSOD}\\
\textbf{Model} & \textbf{Data} & \textbf{$\boldsymbol{F_{max}}$}{$\uparrow$} & {\textbf{MAE}{$\downarrow$}} & $\boldsymbol{S_m}${$\uparrow$} & $\boldsymbol{F_{avg}}\uparrow$ & \textbf{$\boldsymbol{F_{max}}$}{$\uparrow$} & {\textbf{MAE}{$\downarrow$}} & $\boldsymbol{S_m}${$\uparrow$} & $\boldsymbol{F_{avg}}\uparrow$ & \textbf{$\boldsymbol{F_{max}}$}{$\uparrow$} & {\textbf{MAE}{$\downarrow$}} & $\boldsymbol{S_m}${$\uparrow$} & $\boldsymbol{F_{avg}}\uparrow$ & \textbf{$\boldsymbol{F_{max}}$}{$\uparrow$} & {\textbf{MAE}{$\downarrow$}} & $\boldsymbol{S_m}${$\uparrow$} & $\boldsymbol{F_{avg}}\uparrow$ & \textbf{$\boldsymbol{F_{max}}$}{$\uparrow$} & {\textbf{MAE}{$\downarrow$}} & $\boldsymbol{S_m}${$\uparrow$} & $\boldsymbol{F_{avg}}\uparrow$ \\
\midrule
U2Net & orig. & 0.944 & 0.052 & 0.900 & 0.882 & 0.863 & 0.066 & 0.836 & 0.766 & 0.835 & 0.075 & 0.819 & 0.734 & 0.930 & 0.043 & 0.895 & 0.872 & 0.895 & 0.063 & 0.862 & \textbf{0.816}\\
U2Net & ours & \textbf{0.948} & \textbf{0.047} & \textbf{0.908} & \textbf{0.894} & \textbf{0.874} & \textbf{0.061} & \textbf{0.844} & \textbf{0.781} & \textbf{0.848} & \textbf{0.069} & \textbf{0.827} & \textbf{0.746} & \textbf{0.935} & \textbf{0.040} & \textbf{0.900} & \textbf{0.881} & \textbf{0.901} & \textbf{0.062} & \textbf{0.863} & \textbf{0.816}\\
\hline
F3-Net & orig. & 0.955 & 0.035 & 0.924 & 0.917 & 0.899 & 0.038 & 0.887 & 0.841 & 0.831 & \textbf{0.054} & 0.835 & 0.752 & 0.942 & \textbf{0.029} & 0.918 & 0.904 & 0.916 & 0.038 & 0.904 & 0.876\\
F3-Net & ours & \textbf{0.962} & \textbf{0.033} & \textbf{0.929} & \textbf{0.924} & \textbf{0.907} & \textbf{0.037} & \textbf{0.890} & \textbf{0.846} & \textbf{0.850} & 0.055 & \textbf{0.843} & \textbf{0.765} & \textbf{0.947} & \textbf{0.029} & \textbf{0.922} & \textbf{0.909} & \textbf{0.926} & \textbf{0.037} & \textbf{0.907} & \textbf{0.879} \\
\hline
TRACER-4 & orig. & 0.956 & 0.027 & 0.929 & 0.931 & 0.911 & 0.029 & 0.896 & 0.867 & 0.847 & 0.048 & 0.848 & 0.786 & 0.944 & 0.024 & 0.921 & 0.916 & 0.933 & 0.025 & 0.918 & 0.905\\
TRACER-4 & ours & \textbf{0.960} & \textbf{0.026} & \textbf{0.933} & \textbf{0.938} & \textbf{0.918} & \textbf{0.026} & \textbf{0.905} & \textbf{0.883} & \textbf{0.848} & \textbf{0.045} & \textbf{0.853} & \textbf{0.794} & \textbf{0.948} & \textbf{0.022} & \textbf{0.928} & \textbf{0.926} & \textbf{0.936} & \textbf{0.024} & \textbf{0.923} & \textbf{0.911}\\
\bottomrule
\noalign{\smallskip}
\end{tabular}}
\caption{ \textbf{Full Salient Object Detection Evaluation.} Comparison of performance on five benchmark datasets and four metrics. Our method generates samples in a higher than native resolution, improving performance on High-Resolution SOD benachmarks \cite{hrsod}.}
\label{t:saliency_full}
\end{table*}

\begin{figure*}[htb]
\captionsetup[subfigure]{labelformat=empty}
\begin{subfigure}[t]{0.48\textwidth}
 \includegraphics[width=\textwidth]{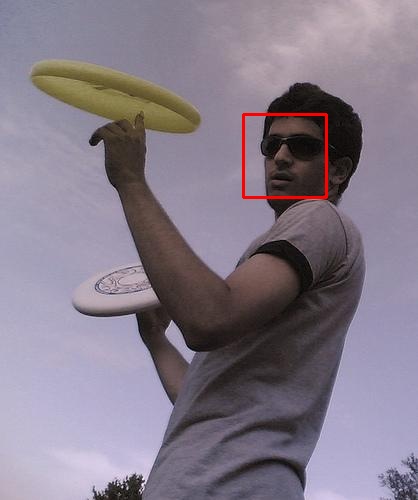}
 \end{subfigure}
\begin{subfigure}[t]{0.48\textwidth}
 \includegraphics[width=\textwidth]{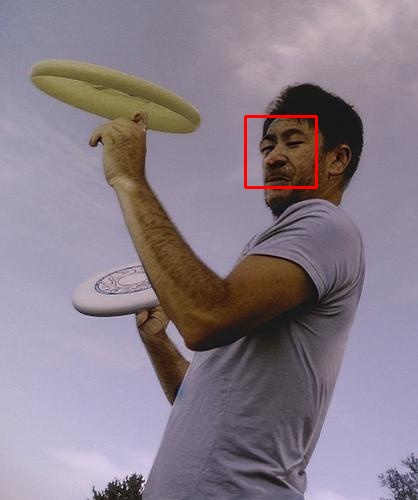}
 \end{subfigure}
\begin{subfigure}[t]{0.48\textwidth}
 \includegraphics[width=\textwidth]{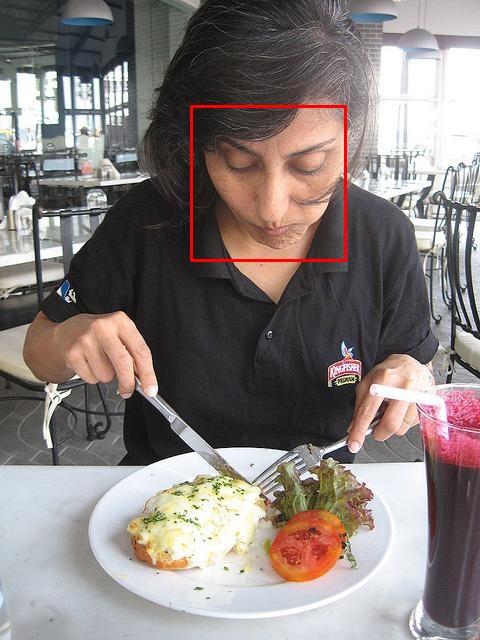}
 \end{subfigure}
\begin{subfigure}[t]{0.48\textwidth}
 \includegraphics[width=\textwidth]{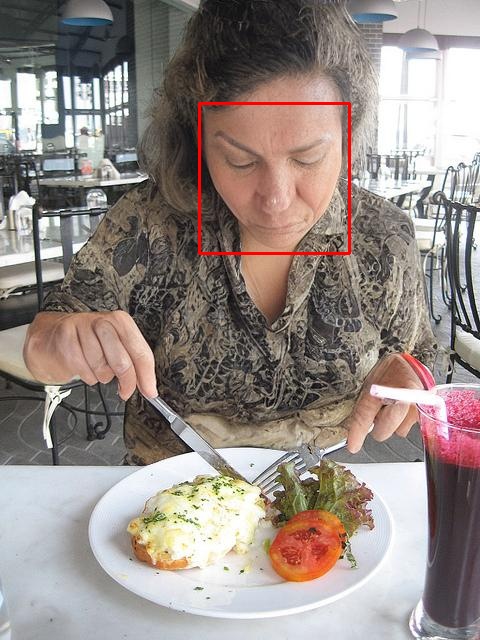}
 \end{subfigure}
\centering
\label{fig:reid_examples}
\end{figure*}

\begin{figure*}[htb]
\captionsetup[subfigure]{labelformat=empty}
\begin{subfigure}[t]{0.47\textwidth}
 \includegraphics[width=\textwidth]{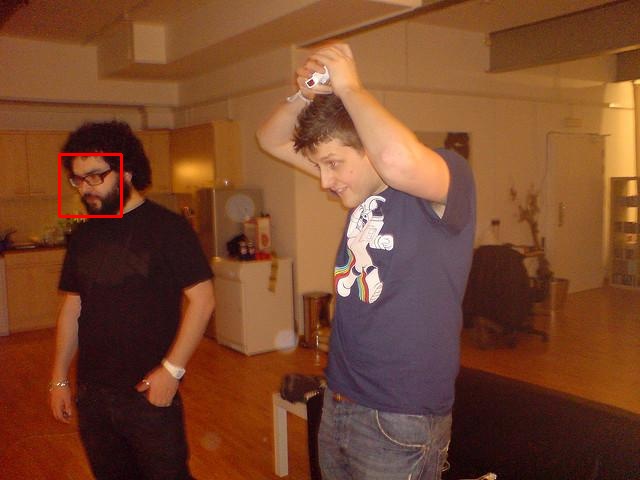}
 \end{subfigure}
\begin{subfigure}[t]{0.47\textwidth}
 \includegraphics[width=\textwidth]{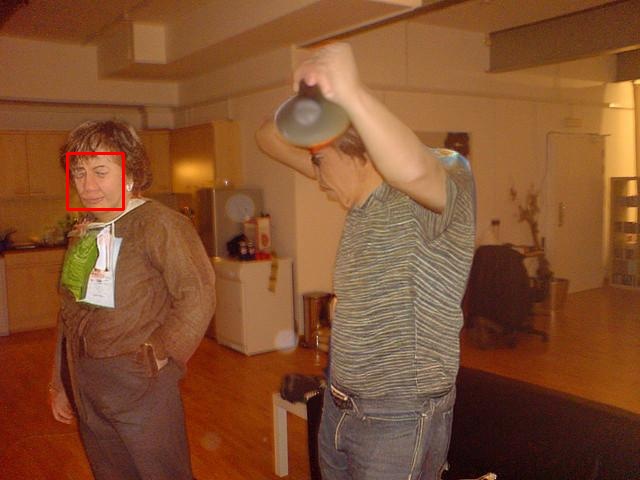}
 \end{subfigure}
 \begin{subfigure}[t]{0.47\textwidth}
 \includegraphics[width=\textwidth]{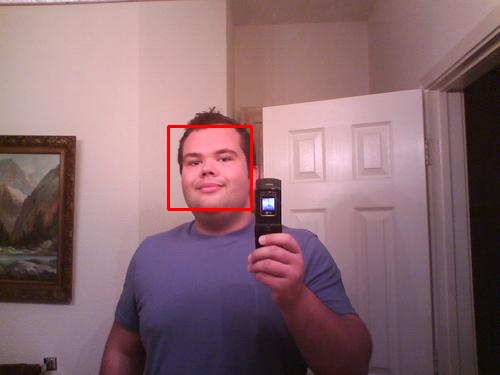}
 \end{subfigure}
\begin{subfigure}[t]{0.47\textwidth}
 \includegraphics[width=\textwidth]{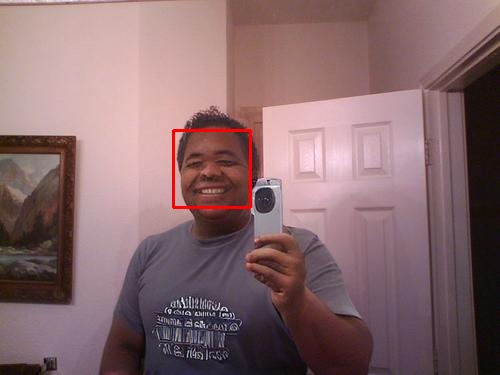}
 \end{subfigure}
 \begin{subfigure}[t]{0.47\textwidth}
 \includegraphics[width=\textwidth]{images/supplementary/reid/train_72239_0.jpg}
 \end{subfigure}
\begin{subfigure}[t]{0.47\textwidth}
 \includegraphics[width=\textwidth]{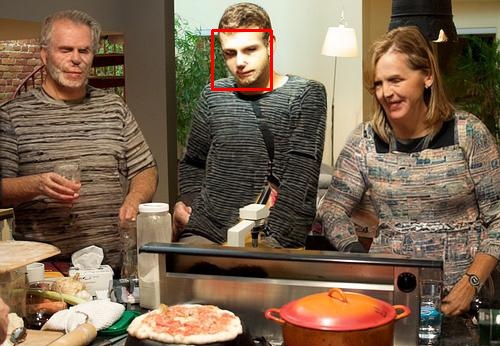}
 \end{subfigure}
\begin{subfigure}[t]{0.47\textwidth}
 \includegraphics[width=\textwidth]{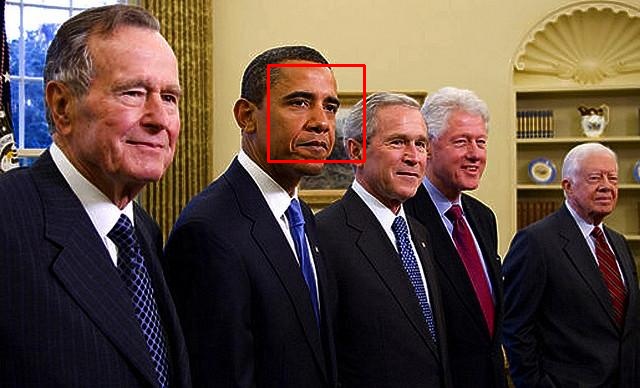}
 \caption{Original}
 \end{subfigure}
\begin{subfigure}[t]{0.47\textwidth}
 \includegraphics[width=\textwidth]{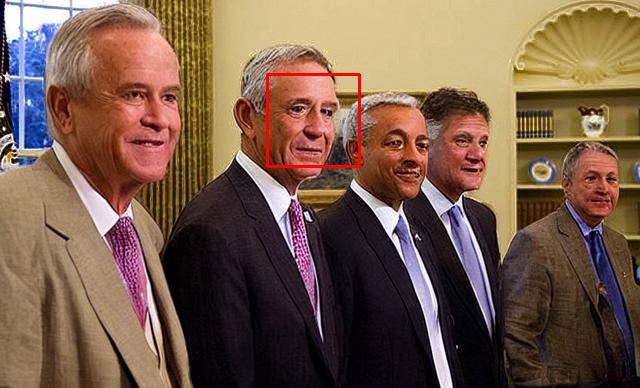}
 \caption{Anonymized}
 \end{subfigure}
\centering
\caption{\textbf{ReIDentified Faces.} Red squares indicate faces that could be matched between the original and generated images. These are likely false positives of the ReID model.}
\label{fig:reid_examples}
\end{figure*}

\clearpage
\bibliographystyle{splncs04}
\bibliography{paper}

\end{document}